\documentclass[10pt]{article} 
\usepackage[preprint]{tmlr}

\usepackage{amsmath,amsfonts,bm}

\def\eqref#1{equation~\ref{#1}}

\def\1{\bm{1}}

\def\vb{{\bm{b}}}

\def\vh{{\bm{h}}}

\def\vr{{\bm{r}}}

\DeclareMathAlphabet{\mathsfit}{\encodingdefault}{\sfdefault}{m}{sl}
\SetMathAlphabet{\mathsfit}{bold}{\encodingdefault}{\sfdefault}{bx}{n}

\usepackage[utf8]{inputenc}
\usepackage{hyperref, url, amsmath, microtype, inconsolata, array, ragged2e, booktabs, multirow, graphicx, amsfonts, lscape, pdflscape, caption, subcaption, mathrsfs, enumitem, amssymb, algorithm, algpseudocode, pifont, xspace, siunitx, nicefrac,xcolor,makecell,csquotes, tikz, diagbox,mathtools,array,tabularx}
\newcommand{\cmark}{\ding{51}}
\newcommand{\xmark}{\ding{55}}
\usetikzlibrary{positioning, shapes.geometric, arrows.meta, calc}

\usepackage[T1]{fontenc}
\usepackage[hang,flushmargin]{footmisc}
\usepackage[normalem]{ulem}
\useunder{\uline}{\ul}{}
\usepackage[most]{tcolorbox}
\algrenewcommand\alglinenumber[1]{\small#1}
\usepackage[export]{adjustbox}

\newcolumntype{M}{>{\raggedright\arraybackslash}p{3cm}}

\newcolumntype{Y}[1]{%
  >{\hsize=#1\hsize\linewidth=\hsize\centering\arraybackslash}X}

\usepackage[super]{nth}
\expandafter\def\expandafter\normalsize\expandafter{%
    \normalsize
    \setlength\abovedisplayskip{7pt}
    \setlength\belowdisplayskip{7pt}
}

\newcommand{\e}{\mathcal{E}}

\newcommand{\cmd}{\textnormal{\textsc{LaSEr-Edit}}\xspace}
\newcommand{\scmd}{\textnormal{\textsc{LaSEr-Edit}}\xspace}
\newcommand{\llmle}{\textnormal{\textsc{LaSEr-LLM Edit}}\xspace}
\newcommand{\mlmle}{\textnormal{\textsc{LaSEr-EBM Edit}}\xspace}

\newcommand{\mlmlere}{\textnormal{\textsc{LaSEr-EBM Edit+LS}}\xspace}
\newcommand{\llmwo}{Plain LLM Edit\xspace}
\newcommand{\llmwoshort}{Plain LLM Edit\xspace}
\newcommand{\llmself}{Self-Locate-LLM Edit\xspace}
\newcommand{\llmselfp}{Self-Parallel-Locate-LLM Edit\xspace}

\newcommand{\toxicsp}{\textnormal{\textsc{ToxicSpans}}}
\newcommand{\inconsp}{\textnormal{\textsc{InconsistentSpans}}}

\newcommand{\roberta}{RoBERTa\xspace}
\newcommand{\gpttwo}{GPT-2~Large\xspace}
\newcommand{\gptthree}{GPT-3.5\xspace}
\newcommand{\gptthreefull}{\texttt{gpt3.5-turbo-0125}\xspace}

\renewcommand{\vb}[1]{\bm{#1}}
\newcommand{\norm}[1]{\left\lVert#1\right\rVert}

\title{\scmd{}: Localized Span-level Error Editing
with Energy-based Localization}

\author{\name Hye Ryung Son \email hyeryung.son@snu.ac.kr \\
      \addr Graduate School of Data Science\\
      Seoul National University
      \AND
      \name Saehee Eom \email seom@gatech.edu \\
      \addr Georgia Institute of Technology\\
      \AND
      \name Mooho Song \email anmh9161@snu.ac.kr\\
      \addr Graduate School of Data Science\\
      Seoul National University
      \AND
      \name Jay-Yoon Lee \email lee.jayyoon@snu.ac.kr\\
      \addr Graduate School of Data Science\\
      Seoul National University}

\def\month{MM}  
\def\year{YYYY} 
\def\openreview{\url{https://openreview.net/forum?id=XXXX}} 

\begin{document}

\maketitle

\begin{abstract}
As large language models (LLMs) are widely adopted in real-world applications, it has become critical to ensure LLMs satisfy safety constraints, such as non-toxicity and logical consistency, as well as task- and situation-specific constraints. Controlling the output through instructions is a simple and tempting approach; however, it remains brittle, is opaque in how it influences model behavior, and thus cannot reliably ensure constraint satisfaction.
Moreover, most recent controlled text generation (CTG) methods require access to the internal components of language models--such as weights or logits--making them incompatible with popular API-based LLMs.
In this work, we propose \cmd{}, a constraint-satisfying text revision method that can be applied to any LLM, black- or white-box. We first find that lightweight, task-specific energy-based models (EBMs) achieve error-localization performance competitive with or even better than that of much larger LLMs, while operating substantially faster. Based on this finding, we propose two variants of text revision methods that incorporate energy-based error localization: \llmle{}, which instructs an LLM to edit text given EBM-predicted error spans, and \mlmle{}, which uses the EBM not only for localization but also for editing by reranking edit candidates. Through experiments in diverse single-constraint control tasks, we show that \llmle{} controls text better than plain LLM-based editing in most of the tasks. We also find that \mlmle{} further improves the control performance of \llmle{} and achieves among the strongest controllability across all tasks. Furthermore, we find that \scmd{}, especially \mlmle, performs well even when multiple constraints are controlled simultaneously.
\end{abstract}

\section{Introduction}

\begin{figure*}[t]
    \centering

    \begin{minipage}[t]{0.46\textwidth}
        \vspace{0pt}
        \centering
        
        \begin{subfigure}[t]{\linewidth}
            \centering
            \includegraphics[width=\linewidth]{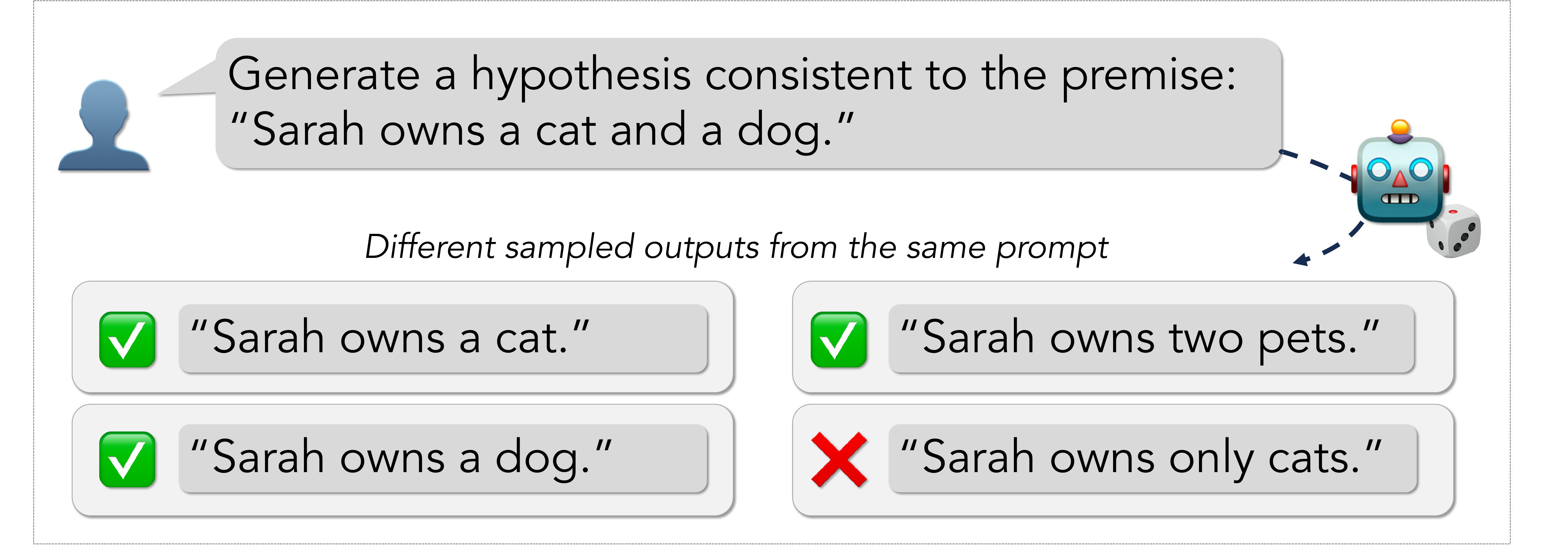}
            \caption{Prompting failure under stochastic generations}
            \label{fig:prompt-nondeterministic}
        \end{subfigure}

        \vspace{1.0em}

        \begin{subfigure}[t]{\linewidth}
            \centering
            \includegraphics[width=\linewidth,
                trim={0 65bp 0 0},
                clip]{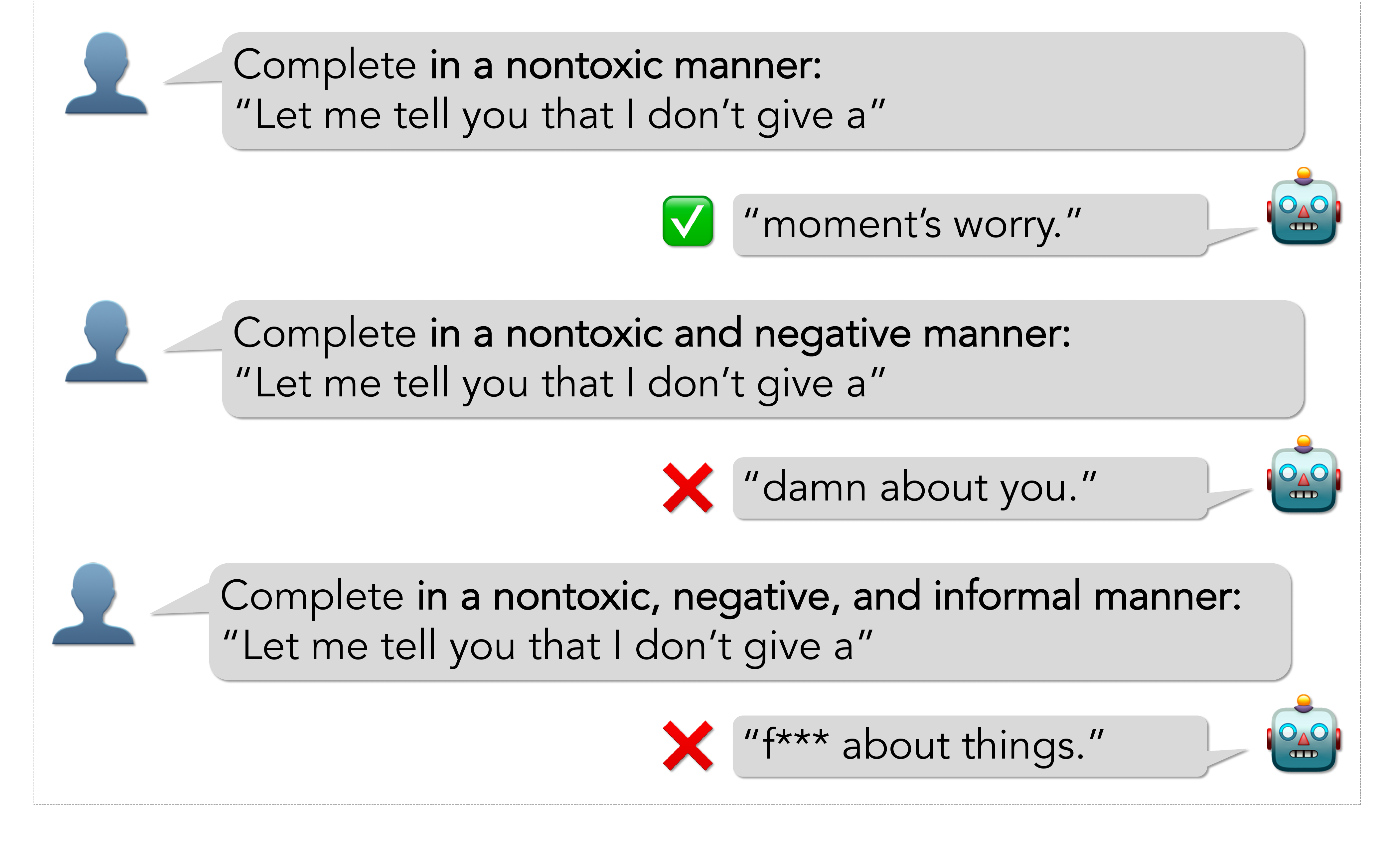}
            \caption{Prompting failure under multi-constraint control}
            \label{fig:prompt-multiattribute}
        \end{subfigure}
    \end{minipage}
    \hfill
    \begin{minipage}[t]{0.52\textwidth}
        \vspace{0pt}
        \centering

        \begin{subfigure}[t]{\linewidth}
            \centering
            \includegraphics[width=\linewidth]{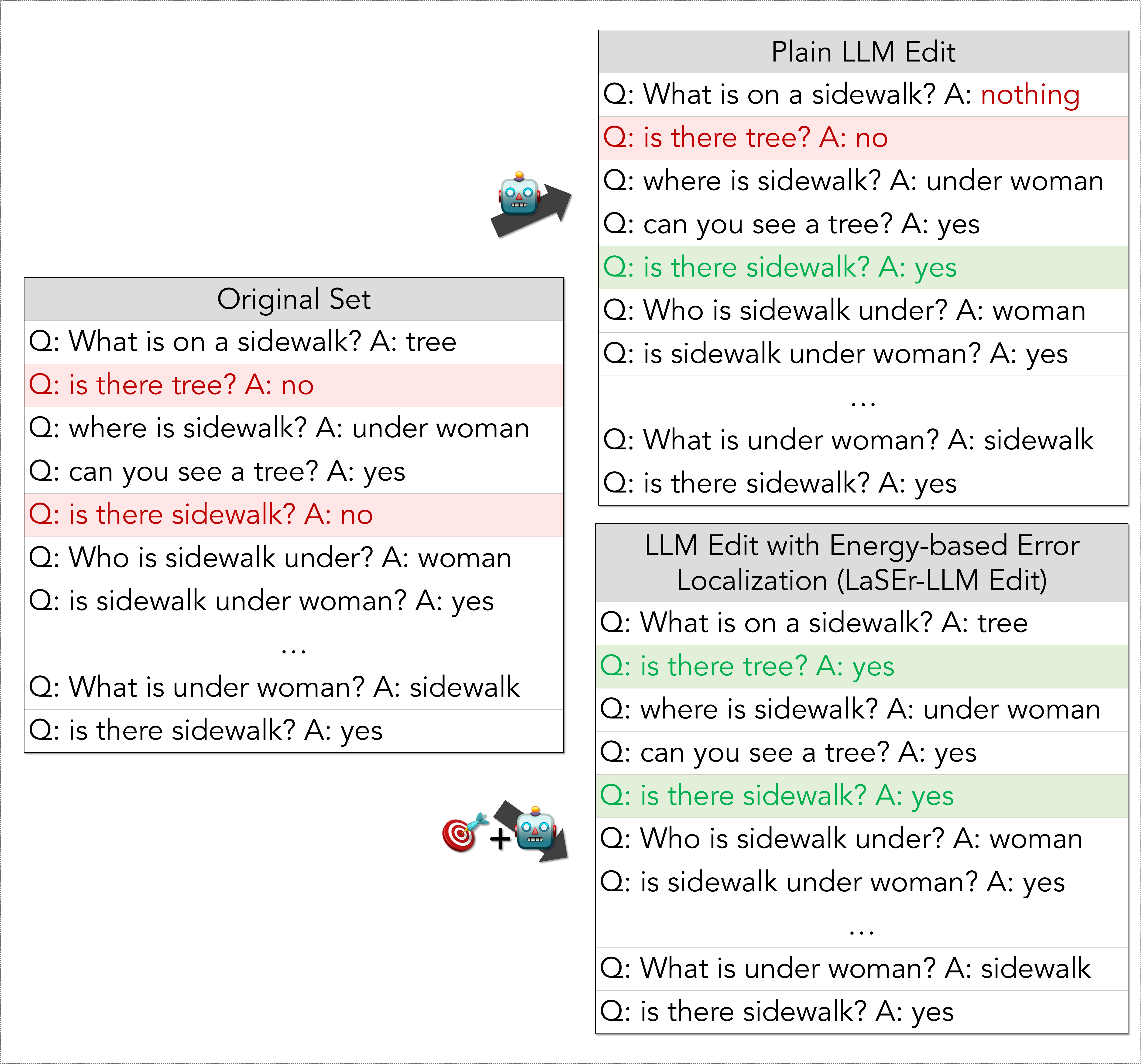}
            \caption{Set-Consistency Enforcement Example}
            \label{fig:llm-edit-better-with-localization}
        \end{subfigure}
    \end{minipage}

    \caption{Common failure modes of prompting-based control (left) and the benefit of energy-based error localization for LLM-based set-consistency enforcement (right). The left panel demonstrates that instruction-based control is non-deterministic due to the stochastic nature of LLM generation and degrades when multiple constraints are combined. The right panel shows that, when provided with error locations predicted by an energy-based model, GPT-5.4 successfully revises all contradiction-causing question-answer pairs—including those missed without error localization—in the Set-LConVQA dataset~\citep{song-etal-2025-introducing}. Full examples are provided in Table~\ref{tab:full-example-figure-1}.}
    \label{fig:motivation}
\end{figure*}

As large language models move from open-ended generation to real-world applications with explicit requirements, the ability to control their outputs has become increasingly important. Prompting provides a convenient interface for steering LLMs, but it is an indirect and brittle form of control. Although prompt engineering can elicit desired behavior without modifying model parameters, the relationship between prompt design and model behavior is not sufficiently systematic to provide fine-grained or deterministic control \citep{sahoo2024systematic,sclar2024quantifying}.  Prior work has shown that LLM behavior can change substantially under superficial prompt variations, including prompt formatting, wording, demonstration choice and order \citep{zhao2021calibrate,lu2022fantastically,chen2023relation,sclar2024quantifying}. More broadly, the stochastic nature of LLM generation makes it difficult for instruction-based control alone to reliably ensure constraint satisfaction (Fig.~\ref{fig:prompt-nondeterministic}). Moreover, adding more constraints to an instruction does not guarantee joint satisfaction (Fig.~\ref{fig:prompt-multiattribute}): models become less reliable as the number of instructions increases \citep{harada2025curse} and exhibit deficiencies on complex instructions composed of multiple constraints \citep{wen2024complexbench,jiang2024followbench}. These findings suggest that prompting alone is often insufficient when applications require reliable, fine-grained control, particularly when constraints must be composed.

Beyond prompting, prior controlled text generation (CTG) methods often steer generation by updating model parameters via training~\citep{gururangan-etal-2020-dont,keskar2019ctrl,zhou2023controlled} or manipulating model internal states such as hidden states, logits, and embeddings at decoding time~\citep{dathathri2020plug,liu-etal-2023-bolt,qin2022cold,kumar-etal-2022-gradient,yang-klein-2021-fudge,kim-etal-2023-critic,pei-etal-2023-preadd}. Although beneficial in that providing additional mechanism than instructions to control LLMs, these methods have limited applicability to popular API-based LLMs that do not provide access to model weights or internal signals.

In addition, many CTG approaches regenerate the entire output when the generated text fails to satisfy the specified constraints. However, regenerated outputs can differ substantially across attempts, leading to two key limitations. First, regeneration may introduce new violations that were not present in the original output. This issue is especially pronounced when multiple constraints are involved, as modifying the output to satisfy one violated constraint may inadvertently cause other constraints to be violated. Second, substantial changes introduced by regeneration make it difficult for humans to track revisions, complicating human-in-the-loop refinement. By contrast, restricting edits to the spans responsible for constraint violations would enable targeted corrections and make iterative efforts toward constraint satisfaction more manageable.

This motivates post-editing as a mechanism for constraint enforcement. Similar to automatic post-editing in machine translation, which corrects initial translations to remove translation errors~\citep{simard-etal-2007-statistical}, we treat constraint enforcement as a revision problem: given an initially generated text, the goal is to correct any constraint violations it contains. In this work, we focus on constraints that can be scored by neural models, particularly energy-based models (EBMs)~\citep{lecun2006ebm}. Because EBMs do not require global normalization for probability scaling, they provide a convenient mechanism for comparing text inputs according to their compatibility with a target constraint. For each constraint, we use a single EBM for two purposes: to identify spans that contribute to constraint violations, using either gradient norms or attention weights, and to rank output candidates by their predicted degree of constraint satisfaction.

In our assessment of error localization performance across various tasks, we find that task-specific, lightweight EBMs achieve localization performance on par or better in comparison to strong LLM baselines while requiring an order of magnitude less execution time. In our evaluation, LLMs exhibit task-dependent fluctuations in localization performance, with no single LLM dominating across all tasks. These results suggest that, for error localization, fine-tuning a small-scale EBM on task-specific training data may provide a more reliable and efficient alternative to relying on the unpredictable localization performance of general-purpose LLMs.

Based on these observations, we propose \scmd{} (\textbf{L}oc\textbf{a}lized \textbf{S}pan-level \textbf{Er}ror \textbf{Edit}ing with Energy-Based Localization), a constraint-satisfying text revision method that edits the text utilizing localized constraint-relevant spans identified by task-specific EBMs. We introduce two editing variants: \llmle{}, which uses an LLM to edit the localized spans, and \mlmle{}, which uses the same EBM used for error localization to guide edit generation. Specifically, in \llmle{}, the LLM is instructed to edit the identified instances, whereas in \mlmle{}, a masked language model proposes token-level replacements for each masked token in the span, and a causal language model and the constraint-specific EBM(s) rank combinations of the token-level candidates to obtain the best span-level edits.
 
In our experiments on toxicity avoidance and set-consistency enforcement, we show, for the first time for constraint-satisfying text revision, that LLM-based editing with EBM-based error localization (\llmle) improves controllability over unlocalized LLM-based editing (Fig~\ref{fig:llm-edit-better-with-localization}; Sec.\ref{sec:single-constraint-exp}). We further find that \mlmle surpasses \llmle in terms of constraint satisfaction. \llmle, on the other hand, provides faster inference and more fluent outputs.

Moreover, we demonstrate that \scmd{} can be extended to multi-constraint settings through an experiment on simultaneous non-toxicity and logical consistency control simply by combining EBMs for each constraint. In this setting, the gap between \scmd{} and unlocalized LLM editing becomes even larger: \mlmle improves the percentage of outputs satisfying both constraints by more than 20 percentage points over \llmwo. These results suggest that localization and energy-based editing are particularly useful when satisfying multiple constraints simultaneously.

More broadly, our results show the value of separating constraint modeling from general-purpose generation. LLMs may not generalize reliably to certain target constraints, such as set-consistency constraints, especially when those constraints involve information or requirements that the model has not previously encountered. Examples include company-specific requirements, domain-specific policies, or arbitrary user-defined constraints. Rather than requiring LLM fine-tuning for each new target, which can be costly and impractical, our approach trains a lightweight encoder-based model for error localization and constraint-enforcing editing. This modular approach enables effective constraint enforcement even for targets previously unseen by LLMs, and it can also be combined with a fine-tuned LLM when available.

\begin{figure*}
    \centering
    \includegraphics[width=\textwidth]{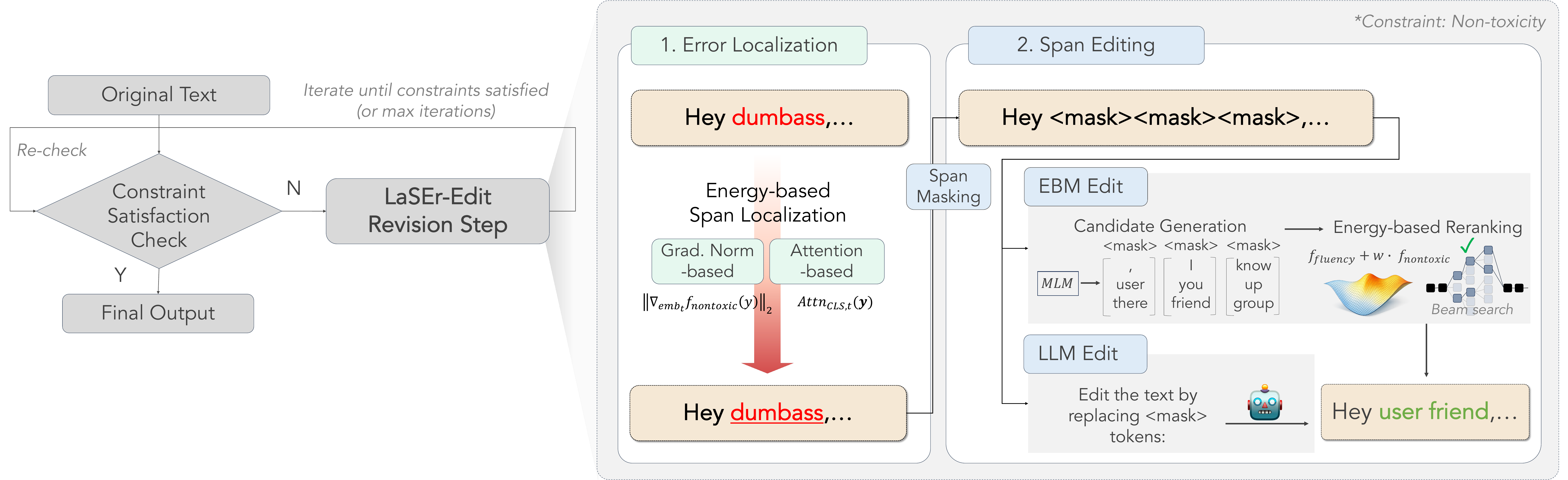}
    \caption{Overview of \scmd{}. Given an initially constraint-violating text, \scmd{} iteratively applies its revision step—energy-based error localization followed by localized span editing—and re-checks constraint satisfaction until the constraint is satisfied. \scmd{} supports two editing variants: \llmle{} and \mlmle{}. In \llmle{}, an LLM is instructed to edit the localized spans, whereas \mlmle{} uses the EBM employed for error localization to perform span editing. Empirically, \mlmle{} provides stronger constraint control, while \llmle{} offers a faster and more fluent alternative.
}
    \label{fig:main_method_wide_laser}
\end{figure*}

Our contributions are as follows:
\begin{itemize}
[itemsep=0pt, topsep=0pt]

\item  We find that task-specific, small-scale fine-tuned EBMs can localize error spans and instances at a level on par with strong LLM baselines, despite having only a fraction of the parameters and operating at much higher speeds. In contrast, LLMs exhibit task-dependent fluctuations in error localization performance, suggesting that training a lightweight task-specific energy model may be more reliable and efficient than relying on a general-purpose LLM for error localization.

\item  We show, for the first time, that feeding the gold error locations improves instructed LLM-based editing for constraint-satisfying text revision. We further show that energy-based localization sometimes yields improvements comparable to those obtained with gold locations.

\item  We propose \scmd{}, a constraint-satisfying text revision method that performs two variants of text edits following energy-based error localization: \llmle and \mlmle. 

\item We show through experiments on three important tasks—toxicity avoidance, pairwise contradiction avoidance, and set-consistency enforcement—that \mlmle provides stronger control, while \llmle runs faster and produces more fluent outputs.
\item Finally, we extend \scmd{} to multi-constraint settings, a more challenging setting where instruction-following has limitations, and find that our localized EBM-based editing outperforms the unlocalized LLM editing baseline for joint constraint satisfaction.
\end{itemize}

\section{\cmd{}}

\subsection{Overview}

\scmd{} is a localized text editing framework that transforms a text sequence $\vb y$ into a revised sequence $\vb y^*$ satisfying a set of constraints $\mathscr{C}=\{c_1,c_2,\ldots,c_n\}$. When $\vb y$ is generated by an LLM, optionally conditioned on a prefix sequence $\vb x$, \scmd{} serves as a post-hoc constraint enforcement method for the LLM.

\scmd{} consists of two stages. First, it performs \textbf{error localization}, where lightweight constraint-specific energy-based models (EBMs) identify spans in $\vb y$ that violate the target constraints. It then performs \textbf{text editing}, modifying only the localized spans using either an LLM or a combination of lightweight models including the same EBMs. These stages may be repeated iteratively to progressively eliminate remaining constraint violations. The complete algorithm is summarized in Algorithm~\ref{alg:locedit} in the appendix.

\subsection{Defining Energy-Based Models}
\subsubsection{Constraint-specific EBMs}
\label{sec:energy-function}
A key component of \scmd{} is a constraint-specific EBM, denoted by $\mathcal{E}_c(\vb y;\theta_c)$, which assigns an energy to a text sequence $\vb y$ with respect to constraint $c$. Lower energy indicates greater compatibility between $\vb y$ and the constraint.

We use each EBM for three purposes. First, it identifies constraint-violating spans during error localization. Second, it scores candidate edits during decoding. Finally, it determines whether a sequence satisfies the corresponding constraint by comparing its energy against a predefined threshold $\epsilon_c$. Specifically, constraint $c$ is considered satisfied if
$\mathcal{E}_c(\vb y;\theta_c) < \epsilon_c$.
When \scmd{} is applied iteratively, this criterion is used to terminate the recursive editing process. It is also used to filter candidates for the final output.

To localize error spans and rank candidate edits, EBMs must provide fine-grained energy signals that enable reliable comparison of candidate samples. For instance, if an EBM can reliably distinguish between two subtly different samples, one erroneous and the other not, its prediction is more likely to depend on the evidence responsible for the error rather than spurious artifacts, thereby supporting both accurate error localization and reliable ranking of the samples. To obtain such fine-grained energy signals, we train each EBM to predict continuous energy values rather than binary labels. Depending on the available supervision, we train the EBM using one of the following objectives:

\begin{itemize}
    \item \textbf{Relaxed cross-entropy.} When soft labels are available (e.g., the average agreement of annotators judging a sample to satisfy a constraint), we optimize a relaxed cross-entropy objective using these labels as supervision.
    
    \item \textbf{Margin-based ranking.} When pairwise preference labels are available, we optimize a margin ranking objective that encourages compatible sequences to receive lower energy than incompatible ones by at least a predefined margin.
\end{itemize}

\subsubsection{Composite Energy Function}
To encourage fluent revisions, we additionally define a fluency energy
$\mathcal{E}_f(\vb y;\xi)
=
-\log p(\vb y\mid\vb x;\xi)$,
where $\xi$ denotes an arbitrary off-the-shelf causal language model. We then combine the fluency energy with the constraint-specific energies to obtain the composite energy 
\begin{equation}
\mathcal{E}(\vb y)
=
w_f\,\mathcal{E}_f(\vb y;\xi)
+
\sum_{c\in\mathscr{C}}
w_c\,\mathcal{E}_c(\vb y;\theta_c),
\label{eq:composite-energy}
\end{equation}
where $w_f$ and $\{w_c\}$ control the trade-off between fluency and constraint satisfaction.
\subsection{Error Localization Methods}
\label{sec:method-locate}
\paragraph{Span Localization Method} For error span localization, we adopt and compare two unsupervised methods proposed in prior work which utilize the \textbf{gradient norm}\citep{li2022oreo} and \textbf{attention weights}\citep{reid-zhong-2021-lewis} of Transformer\citep{vaswani2017attention} encoder-based model. For both methods, we first compute a token-level score for each token $y_t$ in $\vb y$ and select tokens with scores greater than or equal to the average in the sequence. If only a subword of a word is identified, we postprocess to include the other portion such that a whole word is localized.

The \textbf{gradient norm}-based token-level score is calculated as the $L2$ norm of the gradient of energy value with respect to each token's embedding:  \begin{equation}\norm{\nabla_{\textrm{emb}(y_{t})}{\mathcal{E}_{c}(\vb y)}}_2\label{eq:loc-gn}\end{equation} where $\textrm{emb}(y_{t})$ represents the embedding of $y_{t}$. 

The \textbf{attention}-based token-level score is calculated as the maximum value of $j$-th layer's attention weights from the \texttt{CLS} token to $y_{t}$ across multiple heads: 
\begin{equation}\max_{\text{Heads}}\text{Attn}_{0,t}^{j}(\vb y).\label{eq:loc-attn}\end{equation}
\paragraph{Instance Localization Method} 
If the input consists of multiple instances (e.g., sentences or question-answer pairs), we take a two-step approach where we first identify an erroneous instance (sentence or question-answer pair) and then localize text spans within the instance. To express an instance-level score, we simply aggregate token-level scores at the instance-level by taking an average or median. We then choose an instance with maximum instance-level value.

\subsection{Editing Methods}
\label{sec:method-edit}
Given localized errors, we perform editing using either an instructed LLM or a combination of small-scale models including the constraint-specific EBMs. 

\subsubsection{LLM Edit}
In this editing variant, we instruct an LLM to modify only the localized spans in $\vb y$. The localized span information is provided differently depending on the experimental setting, namely the granularity of the localized spans. For phrase- or token-level spans, we mask the corresponding locations in $\vb y$ and provide the masked sequence to the editor LLM. For instance-level spans (i.e., sentences or question--answer pairs), we keep $\vb y$ unchanged and instead specify the indices of the instances to be edited. The exact prompts used in the experiments are listed in Table~\ref{tab:task-prompts-single} in the appendix.

\subsubsection{EBM Edit}
In this editing variant, we first mask the error spans and predict candidate tokens for the masked positions. We then use the same EBMs employed for error localization to rank candidate edits and use beam search to identify the highest-scoring edited output.
The \textsc{EBM Edit} process is summarized in Alg.~\ref{alg:maskinfill}.

\begin{algorithm}[ht]
\caption{\textsc{EBM Edit}}
\scriptsize
\label{alg:maskinfill}

\textbf{Input:} text sequence $\vb y$, localized error spans $\mathcal{S}$, prefix $\vb x$,
MLM, optional LLM for smoothing, energy functions $\e_f$ and $\{\e_c\}$, beam size $n_b$,
number of token candidates $k$, maximum replacement length $m$

\begin{algorithmic}[1]
\State Obtain $\tilde{\vb y}$ by replacing each span in $\mathcal{S}$ with $m$ mask tokens
\State $\mathcal{H} \gets \{$prefix of $\tilde{\vb y}$ before the first masked span$\}$

\For{each masked span in $\tilde{\vb y}$}
    \State $\mathcal{B} \gets \emptyset$
    \For{each hypothesis $\vh \in \mathcal{H}$}
        \State Construct the MLM input by filling previous spans according to $\vh$
        \State $\mathcal{C} \gets$ top-$k$ MLM token candidates for each mask position in the current span \Comment{Candidate Generation}
        \State $\mathcal{H}_0 \gets \{\emptyset\}$
        \State $\mathcal{H}_{\mathrm{cache}} \gets \mathcal{H}_0$
        \For{$i = 1$ to $m$}
            \State $\tilde{\mathcal{H}}_i \gets \emptyset$
            \For{each partial replacement $\vr \in \mathcal{H}_{i-1}$}
                \For{each candidate token $z \in \mathcal{C}_i$}
                    \State $\tilde{\mathcal{H}}_i \gets \tilde{\mathcal{H}}_i \cup \{\vr \circ z\}$
                \EndFor
            \EndFor
            \State $\mathcal{H}_i \gets$ top-$n_b$ replacements in $\tilde{\mathcal{H}}_i$ ranked by the composite energy
            \Comment{Beam Search-based Expansion}
            \State $\mathcal{H}_{\mathrm{cache}} \gets \mathcal{H}_{\mathrm{cache}} \cup \mathcal{H}_i$
        \EndFor

        \State $\mathcal{B}_{\vh} \gets \{\vh \circ \vr : \vr \in \mathcal{H}_{\mathrm{cache}}\}$
        \State $\mathcal{B} \gets \mathcal{B} \cup \mathcal{B}_{\vh}$
    \EndFor
    
    \If{current span is not the last masked span}
        \State Append the text between the current span and the next masked span
        \State $\mathcal{H} \gets$ top-$n_b$ hypotheses ranked by the composite energy
        \Comment{End-of-Span Reranking}
    \Else
        \State Append the remaining suffix 
        \State $\vb y^* \gets$ the hypothesis with the lowest $\e_f(\vb y)$ among those satisfying $\e_c(\vb y)<\epsilon_c$ for all constraints $c$ \Comment{Final Selection}
    \EndIf
\EndFor
\If{LLM smoothing is enabled}
    \State Use an LLM to smooth $\vb y^*$
    \Comment{Optional LLM Smoothing}
\EndIf
\State \Return $\vb y^*$

\end{algorithmic}
\end{algorithm}

\paragraph{Candidate Generation} 

We use an MLM to generate candidate tokens for replacing each mask. This step incorporates bidirectional context into the decoding process.

To allow variable-length replacement, we first replace the current error span (i.e., a contiguous sequence of identified erroneous tokens) with a predefined number $m$ of mask tokens. We then use the fact that encoder-based models such as RoBERTa~\citep{liu2019roberta} are pretrained with a masked language modeling objective: we feed the masked sequence into an off-the-shelf MLM and obtain the likelihoods of candidate tokens for each mask position. 

After ignoring the likelihoods for unmasked tokens and for mask tokens in spans other than the current one, we obtain logits of shape $\mathbb{R}^{|\mathcal{V}| \times m}$. We then select the top-$k$ tokens for each position, resulting in a candidate
index matrix
\[
C \in [|\mathcal{V}|]^{k \times m},
\]
where $\mathcal{V}$ denotes the vocabulary and
$[n] = \{0,1,\ldots,n-1\}$ denotes the set of zero-based indices.

\paragraph{Beam Search-based Expansion} 
Considering all combinations of the $k$ token candidates for each of the
$m$ masked positions would require evaluating $k^m$ span-level candidates, resulting in exponential complexity. To avoid this computational burden, we use a
beam-search-based candidate expansion procedure.

Rather than obtaining expansion candidates from an autoregressive decoder, our procedure expands span-level candidates using token candidates proposed by the MLM, thereby leveraging both the preceding and following context. We rank the resulting candidates using the composite energy defined in Eq.~\ref{eq:composite-energy}, where $w_f$ and $\{w_c\}$ are defined by task. Finally, unlike standard beam search, which retains only the hypotheses of the current length $t$, we retain hypotheses of every length from $0$ to $m$. Formally, let $\mathcal{H}_i$ denote the beam retained after generating $i$ replacement
tokens, with $\mathcal{H}_0=\{\emptyset\}$ , where $\emptyset$ denotes the empty replacement and thus corresponds to deleting the span altogether.
After expanding all $m$ mask positions, we cache the hypotheses
\[
\mathcal{H}_{\mathrm{cache}}
=
\bigcup_{i=0}^{m}\mathcal{H}_i,
\]
where each $\mathcal{H}_i$ contains at most $n_b$ hypotheses, with $n_b$ denoting the beam size. This allows the final reranking stage to compare hypotheses of varying replacement lengths.

\paragraph{End-of-Span Reranking}
When beam search reaches the last mask position of the current span, we perform a final re-ranking over $\mathcal{H}_{\mathrm{cache}}$. Depending on the dataset and constraints, we append the post-context (up to, but not including, the next masked position) to each hypothesis, allowing constraints such as fluency and consistency to account for compatibility with the subsequent context. When re-ranking to select the top $n_b$ hypotheses to carry forward to the next masked span, we use the composite energy in Eq.~\ref{eq:composite-energy}. When re-ranking hypotheses after processing the final masked span, we output the candidate with the lowest $\e_f(\vb y)$ among those satisfying $\e_c(\vb y) < \epsilon_c$ for all constraints $c$.

\paragraph{LLM Smoothing (LS)} We observe in our experiments that text edited by \mlmle{} is sometimes less fluent than the original text, albeit satisfying constraints more strictly. Therefore, we optionally apply a fluency-improvement step using an LLM, which is instructed only to correct grammatical errors and improve sentence flow without modifying the content. We refer to this variant as \mlmlere{}. The prompt used for LLM smoothing is reported in the appendix Table~\ref{tab:llm-smoothing-prompts}.

\section{Experiments and Results}
We organize our experiments in three parts. First, we compare the error localization performance and execution time of task-specific EBMs against a set of strong LLM baselines, assessing whether lightweight EBMs can serve as effective localizers. Second, we evaluate \scmd{} on a diverse set of single-constraint control tasks, namely toxicity avoidance, pairwise contradiction avoidance, and set-consistency enforcement. Finally, we extend \scmd{} to a multi-constraint control setting, targeting both toxicity and logical consistency, to examine whether it remains effective when multiple constraints must be satisfied simultaneously.

\subsection{RQ1. How does a task-specific EBM perform on error localization, in comparison with LLMs?}
\label{sec:error-localization}
\paragraph{Experiment Settings} To evaluate the effectiveness of task-specific EBMs for error localization, we compare them with four LLMs: two reasoning models (GPT-5 mini\footnote{We use the \texttt{gpt-5-mini-2025-08-07} model snapshot.} and Qwen3-8B) and two models evaluated without explicit reasoning tokens (GPT-5.4\footnote{Although GPT-5.4 supports reasoning tokens, we run it without them because of cost constraints. We use the \texttt{gpt-5.4-2026-03-05} model snapshot.} and Qwen2.5-7B-Instruct). We evaluate localization performance and throughput, measured in examples per second, across three tasks. Table~\ref{tab:error-localization-tasks} summarizes the tasks, test datasets, and evaluation metrics. For the LLMs, we use greedy decoding for the open-weight models and default inference settings for the API-based models. The EBM localization implementation details are reported in Table~\ref{tab:hyperparams-locate} in the appendix.

\begin{table}[t]
\centering
\small
\begin{tabular}{p{0.2\linewidth}p{0.34\linewidth}p{0.25\linewidth}p{0.10\linewidth}}
\toprule
\textbf{Task} & \textbf{Goal} & \textbf{Dataset} & \textbf{Metric} \\
\midrule
\textbf{Toxic span detection} 
& Identify the minimal spans in an LLM-generated continuation responsible for toxicity. 
& \toxicsp{} \newline (Appendix~\ref{sec:appendix-span-detection-dataset})
& Recall \\
\midrule
\textbf{Inconsistent span detection} 
& Identify the minimal spans in a hypothesis that contradict a given premise. 
& \inconsp{} \newline (Appendix~\ref{sec:appendix-span-detection-dataset})
& Recall \\
\midrule
\textbf{Inconsistent QA pair detection} 
& Identify the minimal subset of question--answer pairs whose removal restores consistency within a set. 
& 300 inconsistent sets from Set-LConVQA~\citep{song-etal-2025-introducing}
& Exact match \\
\bottomrule
\end{tabular}
\caption{
Error localization tasks and evaluation metrics. For the two span-level tasks, we mainly report recall because missed error spans are more harmful in our editing framework than false positives, which can be preserved during editing. For the instance-level task, we follow \citet{song-etal-2025-introducing} and report exact match.
}
\label{tab:error-localization-tasks}
\end{table}

\begin{figure}[htbp]
    \centering
    \includegraphics[width=\textwidth]{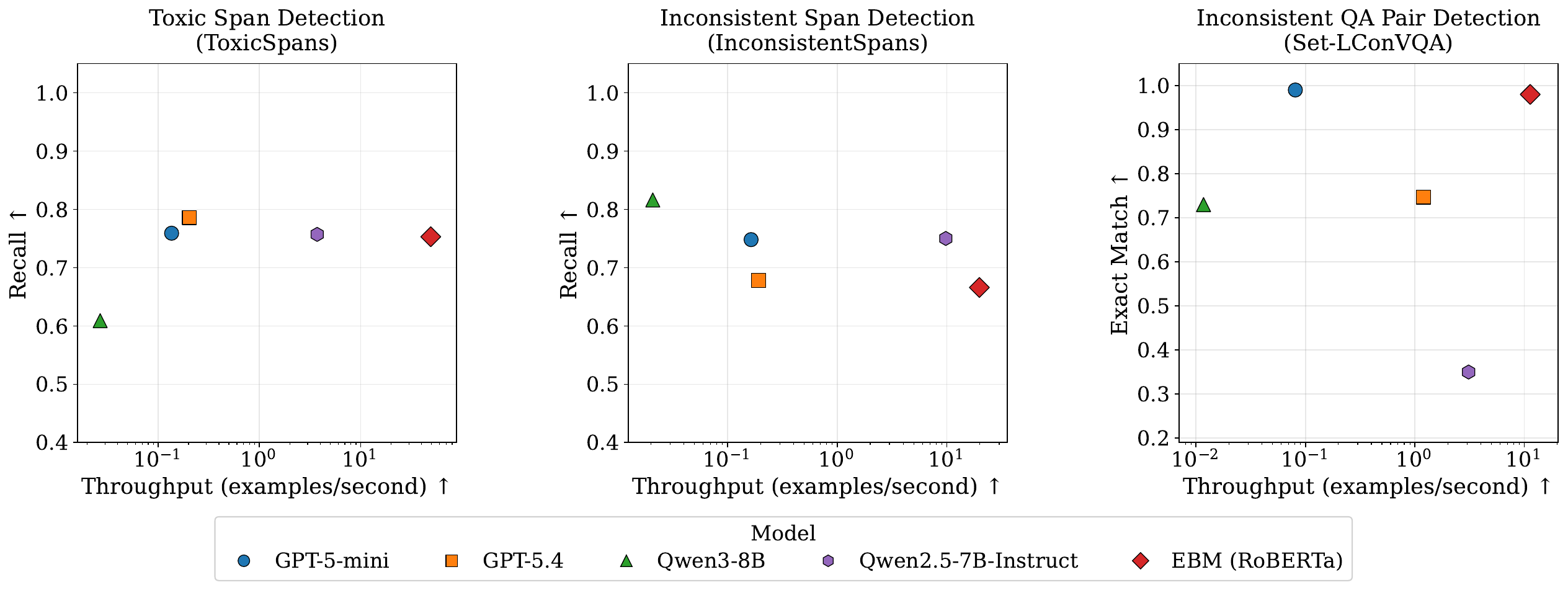}
    \caption{Error localization performance versus throughput across three tasks; dataset names are shown in parentheses. Task-specific EBMs achieve comparable or even better localization accuracy than LLMs while requiring orders of magnitude less execution time. In contrast, each evaluated LLM exhibits noticeable variation in localization accuracy across tasks, suggesting that localization performance depends on the target constraint. This makes relying on a single general-purpose LLM for diverse constraint-control tasks less reliable. Instead, training lightweight task-specific EBMs provides a more scalable and dependable approach to supporting diverse constraints.}
    \label{fig:loc-perf-all}
\end{figure}

\paragraph{Results} Task-specific EBMs consistently achieve localization performance comparable to, and in some cases exceeding, that of much larger LLMs while requiring orders of magnitude less execution time (Figure~\ref{fig:loc-perf-all}). On the Set-LConVQA benchmark, the EBM even achieves nearly perfect localization accuracy, while remaining the fastest model evaluated. Although it trails LLMs by modest margins in inconsistent span detection, we show later in Section~\ref{sec:single-constraint-exp} that the gap can be compensated for by the choice of subsequent editing algorithm, as \mlmle{} using EBM localization outperforms \llmself{} using LLM-based localization (Table~\ref{tab:main-all-macro-avg-tox-nli}). 

The LLMs exhibit larger variability across tasks. No single LLM except GPT-5 mini is competitive across all tasks. While GPT-5 mini performs competitively across all three benchmarks, it comes with a trade-off of substantial computation time. The other models, especially the two open-weight models, show pronounced performance fluctuations. In practice, new application-specific constraints frequently arise after foundation models have been released. Adapting API-based LLMs to such constraints is generally infeasible, while fine-tuning open-weight LLMs remains computationally expensive. In contrast, our framework externalizes constraint modeling to lightweight encoder-based EBMs, which can be trained efficiently for new constraints while maintaining localization quality comparable to substantially larger LLMs.

\subsection{RQ2. Can \scmd{} successfully edit texts to enforce individual constraints?}
\label{sec:single-constraint-exp}
We now examine whether our method \scmd{} can successfully transform a given text to meet a single constraint.
\subsubsection{Task Definitions}
\paragraph{Toxicity Avoidance.} 
In this task, our objective is to detoxify toxic generations produced by a base language model. To construct the test set, we use a black-box LM, \gptthree Turbo (\gptthreefull{}), and generate 10 non-toxic continuations, each up to 150 tokens, for each of 250 non-toxic prompts from RealToxicityPrompts~\citep{gehman-etal-2020-realtoxicityprompts}. In order to conduct a fair comparison among actually edited samples, we report results on 332 continuations whose initial energy value exceeds $\epsilon_c$ we set and are edited by \scmd{} rather than being output as-is. Although $\epsilon_c$ can be set at any level according to the user's need as demonstrated in Sec.~\ref{sec:ab-threshold}, we set $\epsilon_c = -\log0.95$ to attain strong control.
The criterion we used to select $\epsilon_c$ is described in the Appendix~\ref{sec:appendix-hyperparameters}. 

\paragraph{Pairwise Contradiction Avoidance.}
This task evaluates the ability to correct hypotheses generated by a base language model that contradict a given premise. For the test set, we prompt \gptthree{} Turbo to generate 10 consistent hypotheses for each of 1{,}000 premises from the ANLI R2 test set~\citep{nie-etal-2020-adversarial}. Again, for apples-to-apples comparison among actually edited examples, we report results on 3{,}105 generations whose energy values exceed a pre-set $\epsilon_c$ value and are edited by \scmd{}. We set $\epsilon_c = -\log0.99$. 

\paragraph{Set-Consistency Enforcement.}
In this task, we evaluate the ability to edit a set of contradictory texts so that the resulting set becomes consistent. We use two datasets released by \citet{song-etal-2025-introducing}: Set-LConVQA, which consists of sets of question--answer pairs drawn from an existing visual question answering dataset LConVQA~\citep{ray-etal-2019-sunny}, and Set-SNLI, which consists of sets of sentences derived from SNLI~\citep{bowman-etal-2015-large}. We sample 300 inconsistent examples from each dataset. Set-LConVQA is the simpler task, as it only requires editing answer spans, which are typically a single token long. In contrast, Set-SNLI involves more complex forms of contradiction involving more than two sentences and requires active localization of the spans to edit.

\subsubsection{Experiment Settings.}
\label{sec:single-constraint-setting}
\paragraph{Baselines.}
Since \scmd{} is a constraint-enforcing text revision method, we use three editing-based baselines: Mix\&Match, \llmwo{}, and \llmself{}. For LLM-edit-based baselines, we use the same editor LLM, Qwen2.5-7B-Instruct, as \llmle{}.
\begin{itemize}
\item \textbf{Mix\&Match}~\citep{mireshghallah-etal-2022-mix} is a controlled text generation and revision method that utilizes constraint-specific EBMs to conduct Metropolis-Hastings sampling. For text revision tasks, it directly incorporates BERTScore~\citep{zhang2020bertscore} into its energy function to promote content preservation. 
\item \textbf{\llmwo{}} is a baseline where we zero-shot prompt an editor LLM to revise the given text without explicit error localization. 
\item \textbf{\llmself{}} is a method where we prompt an LLM to first localize error spans and then revise the identified spans. As in \llmle{}, we use instance-level localization for set-consistency enforcement tasks.
\end{itemize}

For the toxicity-avoidance and pairwise contradiction-avoidance tasks, where the source texts are LLM-generated outputs, we additionally compare against two controlled text generation (CTG) methods. By design, both methods regenerate the full output when the initial generation fails to satisfy the target constraints.
\begin{itemize}
    \item \textbf{ScoPE}~\citep{yu-etal-2024-controlled} steers a base language model through block-wise editing: it revises each partially generated block and uses the revised text as context for generating the next block. Unlike ours, ScoPE uses editing to guide ongoing generation rather than to revise a completed output post hoc. We include ScoPE because it can be applied to black-box, API-based language models. 
    \item \textbf{MuCoLa}~\citep{kumar-etal-2022-gradient} also employs task-specific EBMs, but to optimize the token embeddings of a base language model. Unlike our method, MuCoLa requires the EBM and the base LM to share an embedding layer and therefore cannot be directly applied to black-box LMs. In our experiments, we evaluate an adapted variant initialized from the source text (MuCoLa (Source-Init)), to evaluate whether its gradient-based inference procedure can be adapted to our setting.
\end{itemize}

\paragraph{\scmd{} Implementation Details.} For the constraint-specific energy-based models used for toxicity and contradiction avoidance, we train RoBERTa-based regression models with a relaxed cross-entropy loss. Further details on EBM training are provided in Appendix~\ref{sec:appendix-energy-training}. For set-consistency enforcement, we use the EBM checkpoints released by \citet{song-etal-2025-introducing}, which were trained with a margin-based loss. We use Qwen2.5-7B-Instruct for \llmle{}, LLM smoothing, and the fluency energy, and RoBERTa-base as the MLM. For error localization, we choose the best performing method and granularity for each setting. In particular, we use instance-level localization for \llmle{} on the set-consistency tasks, because providing span-level localization information degraded performance.

\paragraph{Evaluation Metrics.}  
We evaluate the results in terms of controllability, fluency, content preservation, and speed. First, we report \textbf{control accuracy (Ctrl.)}, defined as the proportion of outputs that satisfy the target constraint according to an external evaluator. For toxicity avoidance, we measure toxicity using the Perspective API\footnote{\url{https://perspectiveapi.com/}}. For contradiction avoidance, we average the class probabilities from three external NLI classifiers\footnote{\path{ynie/roberta-large-snli_mnli_fever_anli_R1_R2_R3-nli}, \path{cross-encoder/nli-roberta-base}, \path{cross-encoder/nli-deberta-v3-base}; all hosted on \url{https://huggingface.co}.} and classify an output as consistent if it is either entailment or neutral. For set-consistency enforcement, we instruct GPT-5 mini with 5-shot CoT prompting to verify set-consistency.\footnote{
As a complementary analysis, we also provide EBM-based evaluation results in Table~\ref{tab:main-sc-energy-alt} in the appendix, given the finding of \citet{song-etal-2025-introducing} that EBMs are more accurate than LLMs at set-consistency verification. We use GPT-5 mini as the primary evaluator because it is independent of all compared methods, whereas EBMs are used in \scmd{}.} For toxicity avoidance task, because control accuracy alone does not provide sufficient resolution to distinguish among methods, we additionally include average toxicity (\textit{Avg. Tox.}) as measured by Perspective API. Second, we report \textbf{perplexity (PPL)} as a fluency metric, measured with Qwen2.5-14B~\citep{yang2024qwen2}\footnote{We report corpus-level perplexity, computed from the aggregate negative log-likelihood over all tokens, rather than averaging per-sample perplexities, which can have high variance.}. We also report \textbf{$\Delta$PPL}, the absolute difference between the perplexity of the original texts and that of each method’s outputs, to measure deviations from the fluency level of the original text. Third, we report \textbf{BERTScore (Prsv.)} between the original and edited texts to measure semantic preservation. Finally, we report execution speed in \textbf{decoded tokens per second (Speed)}, computed as the number of tokens in the final output divided by the end-to-end wall-clock time in seconds. All local speed measurements are conducted under the same hardware and batch-size conditions: a single NVIDIA RTX A6000 GPU with a batch size of 1.

\subsubsection{Results}

\begin{table*}
\centering
\scriptsize
\setlength\tabcolsep{1pt}
\begin{tabular}{@{}lccccc|cccc|cccc@{}}
\toprule
& \multicolumn{5}{c}{\textbf{Toxicity Avoidance}}
& \multicolumn{4}{c}{\textbf{Contradiction Avoidance}}
& \multicolumn{4}{c}{\textbf{Macro Average}} \\
\cmidrule{2-14}
& \begin{tabular}[c]{@{}c@{}}Ctrl.\\$\uparrow$\end{tabular}
& \begin{tabular}[c]{@{}c@{}}Avg \\Tox.$\downarrow$\end{tabular}
& \begin{tabular}[c]{@{}c@{}}PPL$\downarrow$\\($\Delta$PPL$\downarrow$)\end{tabular}
& \begin{tabular}[c]{@{}c@{}}Prsv.\\$\uparrow$\end{tabular}
& \begin{tabular}[c]{@{}c@{}}Speed\\$\uparrow$\end{tabular}
& \begin{tabular}[c]{@{}c@{}}Ctrl.\\$\uparrow$\end{tabular}
& \begin{tabular}[c]{@{}c@{}}PPL$\downarrow$\\($\Delta$PPL$\downarrow$)\end{tabular}
& \begin{tabular}[c]{@{}c@{}}Prsv.\\$\uparrow$\end{tabular}
& \begin{tabular}[c]{@{}c@{}}Speed\\$\uparrow$\end{tabular}

& \begin{tabular}[c]{@{}c@{}}Ctrl.\\$\uparrow$\end{tabular}
& \begin{tabular}[c]{@{}c@{}}PPL$\downarrow$\\($\Delta$PPL$\downarrow$)\end{tabular}
& \begin{tabular}[c]{@{}c@{}}Prsv.\\$\uparrow$\end{tabular}
& \begin{tabular}[c]{@{}c@{}}Speed\\$\uparrow$\end{tabular}
\\ \midrule
ScoPE\textsuperscript{\dag} & 0.994 & \uline{0.079} & 7.95 (2.70) & 0.19 & 12.38 & 0.894&  \hspace{0.5em}\textbf{9.83} (3.10) & 0.19 & \textbf{18.54} & 0.944 & \hspace{0.5em}\textbf{8.89} (2.90) & 0.19 & \textbf{15.46} \\
MuCoLa (Source-Init)\textsuperscript{\dag\ddag} & 0.961 & 0.143 & \textbf{5.07} (\uline{0.17}) & 0.82 & \hspace{0.5em}1.27 & 0.872 & 15.74 (2.81) & 0.55 & \hspace{0.5em}0.50 
& 0.917 & 10.41 (1.49) & 0.69 & \hspace{0.5em}0.89 \\
Mix\&Match
& 0.991 & 0.118 & 6.04 (0.80) & \textbf{0.94} & \hspace{0.5em}0.06
& \textbf{0.970} & 12.09 (\uline{0.85}) & 0.85 & \hspace{0.5em}0.21
& \textbf{0.980} & \hspace{0.5em}9.07 (\uline{0.82}) & \textbf{0.89} & \hspace{0.5em}0.13 \\
\llmwo
& \textbf{1.000} & 0.139 & 5.97 (0.73) & 0.71 & \uline{13.73}
& 0.942 & \uline{11.86} (1.07) & 0.79 & \uline{12.38}
& 0.971 & \hspace{0.5em}\uline{8.92} (0.90) & 0.75 & \uline{13.06} \\
\llmself & \textbf{1.000} & 0.137 & \uline{5.29} (\textbf{0.04}) & 0.78 & 12.35 & 0.921 & 13.37 (\textbf{0.44}) & \textbf{0.90} & \hspace{0.5em}7.67 & 0.961 & \hspace{0.5em}9.33 (\textbf{0.24}) & \uline{0.84} & 10.01 \\
\midrule
\llmle
& \textbf{1.000} & 0.109 & 6.11 (0.86) & 0.78 & \textbf{13.99}
& 0.900 & 14.07 (1.13) & \uline{0.87} & 11.15
& 0.950 & 10.09 (1.00) & 0.82 & 12.57 \\
\mlmle
& 0.997 & \textbf{0.067} & 6.49 (1.25) & \uline{0.85} & \hspace{0.5em}3.44
& \uline{0.952} & 17.15 (4.21) & 0.71 & \hspace{0.5em}3.81
& \uline{0.974} & 11.82 (2.73) & 0.78 & \hspace{0.5em}3.63 \\
\mlmlere
& 0.997 & \uline{0.079} & 6.36 (1.12)  & 0.81 & \hspace{0.5em}3.21
& 0.949 & 14.71 (1.78) & 0.72 & \hspace{0.5em}3.10
& 0.973 & 10.54 (1.45) & 0.76 & \hspace{0.5em}3.15 \\
\bottomrule
\end{tabular}%
\caption{Experimental results on toxicity avoidance and contradiction avoidance. We report constraint-control accuracy (\textit{Ctrl.}), defined as the proportion of final outputs that satisfy the target constraint according to external evaluators; perplexity (\textit{PPL}) and its absolute difference from that of the original text (\textit{$\Delta$PPL}, shown in parentheses); BERTScore relative to the original text (\textit{Prsv.}); and decoding throughput in tokens per second (\textit{Speed}). For toxicity avoidance, we additionally report average toxicity (\textit{Avg. Tox.}). The original texts have perplexities of 5.24 and 12.93 for toxicity avoidance and contradiction avoidance, respectively. Methods marked with \textsuperscript{\dag} are regeneration-based controlled generation methods; among them, the method additionally marked with \textsuperscript{\ddag} is incompatible with black-box language models. For MuCoLa, we evaluate an adapted variant initialized from the source text to assess whether its gradient-based inference procedure can be applied in our revision setting. Across the evaluated tasks, \mlmle{} achieves among the strongest controllability, whereas \llmle{} offers competitive controllability with greater speed and fluency. Generation examples are provided in Tables~\ref{tab:generation-examples-toxicity} and~\ref{tab:generation-examples-nli} in the appendix.
}
\label{tab:main-all-macro-avg-tox-nli}
\end{table*}

\begin{table*}
\centering
\scriptsize
\setlength\tabcolsep{1pt}
\begin{tabular}{@{}lcccc|cccc|cccc@{}}
\toprule
& \multicolumn{8}{c}{\begin{tabular}[c]{@{}c@{}}\textbf{Set-Consistency Enforcement}\end{tabular}}\\
& \multicolumn{4}{c}{\begin{tabular}[c]{@{}c@{}}{Set-LConVQA}\end{tabular}}
& \multicolumn{4}{c}{\begin{tabular}[c]{@{}c@{}}{Set-SNLI}\end{tabular}}
& \multicolumn{4}{c}{\textbf{Macro Average}} \\
\cmidrule{2-13}
& \begin{tabular}[c]{@{}c@{}}Ctrl.\\$\uparrow$\end{tabular}
& \begin{tabular}[c]{@{}c@{}}PPL$\downarrow$\\($\Delta$PPL$\downarrow$)\end{tabular}
& \begin{tabular}[c]{@{}c@{}}Prsv.\\$\uparrow$\end{tabular}
& \begin{tabular}[c]{@{}c@{}}Speed\\$\uparrow$\end{tabular}
& \begin{tabular}[c]{@{}c@{}}Ctrl.\\$\uparrow$\end{tabular}
& \begin{tabular}[c]{@{}c@{}}PPL$\downarrow$\\($\Delta$PPL$\downarrow$)\end{tabular}
& \begin{tabular}[c]{@{}c@{}}Prsv.\\$\uparrow$\end{tabular}
& \begin{tabular}[c]{@{}c@{}}Speed\\$\uparrow$\end{tabular}
& \begin{tabular}[c]{@{}c@{}}Ctrl.\\$\uparrow$\end{tabular}
& \begin{tabular}[c]{@{}c@{}}PPL$\downarrow$\\($\Delta$PPL$\downarrow$)\end{tabular}
& \begin{tabular}[c]{@{}c@{}}Prsv.\\$\uparrow$\end{tabular}
& \begin{tabular}[c]{@{}c@{}}Speed\\$\uparrow$\end{tabular}
\\ \midrule
Original Text
& 0.003 & 4.31 (0.00) & 1.00 & -
& 0.007 & 3.88 (0.00) & 1.00 & -
& 0.01 & 4.10 (0.00) & 1.00 & - \\
\midrule
Mix\&Match
& - & - & - & \hspace{1.0em}0.01
& - & - & - & \hspace{0.5em}0.01
& - & - & - & \hspace{1.0em}0.01 \\
\llmwo
& 0.327 & \uline{4.11} (0.21) & 0.93 & \hspace{0.5em}38.69
& 0.107 & \textbf{3.88} (\textbf{0.00}) & 0.73 & \textbf{38.76}
& 0.22 & \uline{3.99} (\textbf{0.10})  & 0.83 & \hspace{0.5em}38.72 \\
\llmself & 0.363 & 4.24 (\textbf{0.07}) & 0.96 & \hspace{0.5em}35.90 & 0.180 & \uline{4.04} (\uline{0.15}) & 0.72 & 37.00 & 0.28 & 4.14 (\uline{0.11}) & 0.84 & \hspace{0.5em}36.45 \\
\midrule
\llmle
& 0.453 & 4.18 (\uline{0.13}) & \uline{0.97} & \hspace{0.5em}38.29
& 0.240 & 4.12 (0.23) & 0.70 & \uline{38.10}
& 0.35 & 4.15 (0.18) & 0.83 & \hspace{0.5em}38.20 \\
\mlmle
& \textbf{0.970} & 4.18 (\uline{0.13}) & \textbf{0.98} & \textbf{465.87}
& \uline{0.413} & 4.36 (0.47) & \textbf{0.91} & 18.29
& \textbf{0.69} & 4.27 (0.30) & \textbf{0.94} & \textbf{242.08} \\
\mlmlere
& \uline{0.930} & \textbf{3.58} (0.73) & 0.85 & \uline{451.36}
& \textbf{0.417} & 4.31 (0.43) & \uline{0.90} & 16.77
& \uline{0.67} & \textbf{3.95} (0.58) & \uline{0.87} & \uline{234.06} \\
\bottomrule
\end{tabular}%
\caption{Experimental results for applying control methods to set-consistency enforcement task on two datasets: Set-LConVQA and Set-SNLI dataset~\citep{song-etal-2025-introducing}. For Mix\&Match, the results are omitted because a full evaluation was estimated
to require approximately 54.5 days on a single NVIDIA A6000 GPU. \mlmle{} shows substantially higher control accuracy than all LLM-based editing methods, while \llmle{} outperforms \llmwo{} or \llmself{}. We observe a similar trend when using our EBM rather than an external LLM (GPT-5 mini) as the control accuracy evaluator (Table~\ref{tab:main-sc-energy-alt}). Refer to Tables~\ref{tab:generation-examples-lconvqa} and \ref{tab:generation-examples-set-nli} in the appendix for generation examples.
}
\label{tab:main-all-macro-avg-sc}
\end{table*}

\paragraph{\scmd{} is on par with or better than the baselines in control strength.}
As shown in Tables~\ref{tab:main-all-macro-avg-tox-nli} and~\ref{tab:main-all-macro-avg-sc}, \mlmle{} ranks first or among the top-performing methods across all tasks, demonstrating the effectiveness of localized editing with energy-based reranking for constraint satisfaction. Although Mix\&Match achieves higher control accuracy on contradiction avoidance, it is approximately 18 times
slower than \mlmle{}. The advantage of \mlmle{} is especially pronounced on the set-consistency tasks, suggesting that using a task-specific EBM for editing is more reliable when the editor LLM is not well adapted to the target constraint. \llmle{} also outperforms the baselines on all tasks except contradiction avoidance. On most tasks, it surpasses not only \llmwo{} but also \llmself{}, indicating that task-specific EBMs provide useful error-localization guidance. Moreover, on Set-LConVQA, \llmle{} achieves a control accuracy of 0.453, closely approaching the 0.477 achieved by LLM Edit with oracle error locations.
 

\paragraph{Among the two variants, \mlmle{} excels in controllability, whereas \llmle{} is generally faster and more fluent.}
Across all tasks, \mlmle{} achieves stronger control than \llmle{}. We also note that, as shown in Sec.~\ref{sec:ab-threshold}, the threshold hyperparameter in \mlmle{} allows its control strength to be adjusted more finely than in \llmle{}. However, this stronger controllability comes with a trade-off in speed and fluency. \mlmle{} is 2.1 to 4.1 times slower than \llmle{} on all tasks except set-consistency enforcement on Set-LConVQA, where it is 12 times faster. We attribute this trade-off to the bottleneck of edit candidate evaluation, which is alleviated only in special cases.\footnote{Based on our analysis, \mlmle{} has time complexity $\mathcal{O}\left(\frac{m \cdot l \cdot k \cdot n_b}{B} \cdot N\right)$, whereas \llmle{} has time complexity $\mathcal{O}(L \cdot N)$. Thus, \mlmle{} is faster than \llmle{} only when $\frac{m \cdot l \cdot k \cdot n_b}{B} < L$. Here, $m$, $l$, $k$, and $n_b$ are defined in Alg.~\ref{alg:maskinfill}, and $B$ denotes the batch size used to evaluate candidate edits with the EBM in parallel.} \mlmle{} is also less fluent than \llmle{}, which is to a degree mitigated by additional LLM Smoothing. Thus, the two variants support different use cases: \mlmle{} favors stronger and more fine-grained control, whereas \llmle{} favors faster and more fluent revision.

\paragraph{Applying LLM Smoothing improves the fluency of \mlmle{} with minimal impact on control performance.}
We observe that \mlmle{} generally yields lower fluency than the baselines, as shown by higher $PPL$ and $\Delta PPL$. Our qualitative analysis further shows that it can occasionally produce unnatural or broken sentences. Since LLM-based editing typically produces more fluent outputs, we mitigate this limitation by applying a final LLM-based rephrasing step, which we call LLM Smoothing (LS), that modifies the sentence only to improve fluency. Across all experiments, we observe that \mlmlere{} reduces perplexity while minimally degrading or even improving control accuracy.

\begin{table}[t]
\centering
\small
\setlength{\tabcolsep}{2pt}
\begin{tabular}{lcccccc}
\toprule
& \multicolumn{4}{c}{\textbf{Control}} & \textbf{Fluency} & \textbf{Prsv.} \\ \cmidrule{2-7}
\textbf{Method} 
& \makecell{\textbf{Both}\\\textbf{Sat.}(\%) $\uparrow$} 
& \makecell{\textbf{Only}\\\textbf{Non-Tox.}(\%)} 
& \makecell{\textbf{Only}\\\textbf{Cons.}(\%)} 
& \makecell{\textbf{Neither}\\\textbf{Sat.}(\%) $\downarrow$} 
& \makecell{\textbf{PPL} $\downarrow$\\ (\textbf{$\Delta$}\textbf{PPL} $\downarrow$)} 
& \makecell{\textbf{BERT}\\\textbf{-Score $\uparrow$}} \\
\midrule
Original Text & \hspace{0.5em}0.00 & \hspace{0.5em}0.00 & 0.00 & 100.00 & 53.6 (\hspace{0.5em}0.0)  & 1.00 \\
\midrule
\llmwo 
& 66.73 & 32.38 & 0.18 & \hspace{0.5em}0.72 & \textbf{13.1} (40.5)  & 0.45 \\
\llmself & 61.72 & 38.10 & 0.00 & \hspace{0.5em}\textbf{0.18} & 29.4 (24.2) & \textbf{0.51} \\
\llmselfp & 79.96 & 19.68 & 0.18 & \hspace{0.5em}\textbf{0.18} & \uline{22.7} (31.0) & 0.43 \\
\midrule
\llmle 
& 69.95 & 29.87 & 0.00 & \hspace{0.5em}\textbf{0.18} & 26.3 (27.3) & \uline{0.46} \\ 
\mlmle 
& \textbf{86.76} & 10.55 & 2.33 & \hspace{0.5em}0.36 & 36.7 (\textbf{16.9}) & 0.39 \\
\mlmlere 
& \uline{86.05} & 11.63 & 1.79 & \hspace{0.5em}0.54 & 33.9 (\uline{19.7}) & 0.40 \\
\bottomrule
\end{tabular}
\caption{Results for joint non-toxicity and consistency control. \mlmle{} substantially outperforms all baselines and \llmle{}, demonstrating the effectiveness of energy-guided editing in the multi-constraint setting. \llmle{} also improves over \llmwo{}, suggesting that energy-based error localization can enhance LLM-based editing under multiple constraints.}
\label{tab:joint-nontox-consistency}
\end{table}

\subsection{RQ3. Can \scmd{} enforce multiple constraints simultaneously?}
\label{sec:multi-constraint}
\subsubsection{Experiment Settings}
\paragraph{Task Definition.} We evaluate whether each method can revise texts that are both toxic and contradictory to be nontoxic and consistent. For the test set, we sample 500 contradictory premise--hypothesis pairs from the SNLI and ANLI test splits and use Qwen3-8B to rewrite each hypothesis into two toxic variants. We then filter 561 premise--hypothesis pairs using the external toxicity and NLI classifiers described in Section~\ref{sec:single-constraint-setting} in which the hypothesis is both toxic and contradictory to the premise. Given each pair, the goal is to revise the hypothesis so that it is non-toxic and consistent with the premise.

\paragraph{Baselines.}
We compare against \llmwo{}, \llmself{}, and \llmselfp{}. For \llmself{}, we prompt the editor LLM once to jointly localize all spans that violate at least one of the constraints. By contrast, \llmselfp{} prompts the LLM separately for each constraint and takes the union of the resulting spans. We include \llmselfp{} to align its localization procedure with that of \llmle{}, which likewise performs constraint-specific localization. We use Qwen2.5-7B-Instruct as the editor LLM. The error-localization and editing prompts are provided in Tables~\ref{tab:task-prompts-localization} and~\ref{tab:task-prompts-multi}, respectively, in the appendix.

\paragraph{Extending \scmd{} to multiple constraints.}
For error localization, we independently identify spans using each
constraint-specific EBM and take their union:
\[
\mathcal{S}(\vb y)
=
\bigcup_{c \in \mathcal{C}} \mathcal{S}_c(\vb y),
\]
where $\mathcal{S}_c(\vb y)$ denotes the spans localized for constraint $c$.
During reranking in \mlmle{}, we use the composite energy defined in
Eq.~\ref{eq:composite-energy}. We regard a candidate as satisfying all
constraints if
\[
\vb y \in \mathcal{F}
\coloneqq
\left\{
\vb y :
\mathcal{E}_c(\vb y) < \epsilon_c
\;\; \forall c \in \mathcal{C}
\right\}.
\]
For implementation, we use the same models as in
Sec.~\ref{sec:single-constraint-exp} and the LLM editing prompts provided in
Table~\ref{tab:task-prompts-multi}.

\paragraph{Evaluation Metrics.}
In addition to PPL, $\Delta$PPL, and BERTScore, we report the percentages of samples that satisfy both constraints, satisfy only the non-toxicity constraint, satisfy only the consistency constraint, or satisfy neither constraint. 

\subsubsection{Results}
\paragraph{Even if an editor LLM performs well on individual constraints, it can struggle to enforce them jointly.}
In Sec.~\ref{sec:single-constraint-exp}, we observe that \llmwo{} achieves control accuracies of up to 100\% and 94.2\% for toxicity avoidance and contradiction avoidance, respectively. However, when required to satisfy both constraints simultaneously, its success rate drops to 66.73\% as shown in Table~\ref{tab:joint-nontox-consistency}. Among the failure cases, the case where it only satisfies toxicity constraint is most prevalent. The lower performance of \llmself{} further suggests that LLMs also struggle to localize error spans when multiple constraints must be considered jointly.

\paragraph{\mlmle{} achieves the strongest joint constraint control, highlighting the benefit of energy-guided revision.}
Among all methods, \mlmle{} achieves the highest joint constraint-satisfaction rate, reaching 86.76\%. This substantially exceeds the performance of the baselines and \llmle{}, whose success rates range from 61.72\% to 79.96\%. The result suggests that directly guiding revision with constraint-specific energy models provides stronger and more reliable control than relying solely on an instructed LLM editor. 

\paragraph{Providing localized error spans can improve an LLM editor's adherence to multiple constraints.}
Both \llmle{} and \llmselfp{} improve the joint constraint-satisfaction rate over \llmwo{}, indicating that explicit error localization---whether EBM- or LLM-based---can facilitate LLM editing. By contrast, \llmself{} performs worse than \llmwo{}, suggesting that the benefits of localization depend on its quality and strategy. We also observe that \llmselfp{} outperforms \llmle{}. This may be partly attributable to the stronger single-constraint localization performance of Qwen2.5-7B-Instruct relative to the task-specific EBMs.

\section{Analyses}

\subsection{The effect of adjusting \texorpdfstring{$\epsilon_c$}{epsilon} on \mlmle}
\label{sec:ab-threshold}

\begin{figure}[htbp]
    \centering
    \begin{subfigure}[b]{0.48\textwidth}
        \centering
        \includegraphics[width=\textwidth]{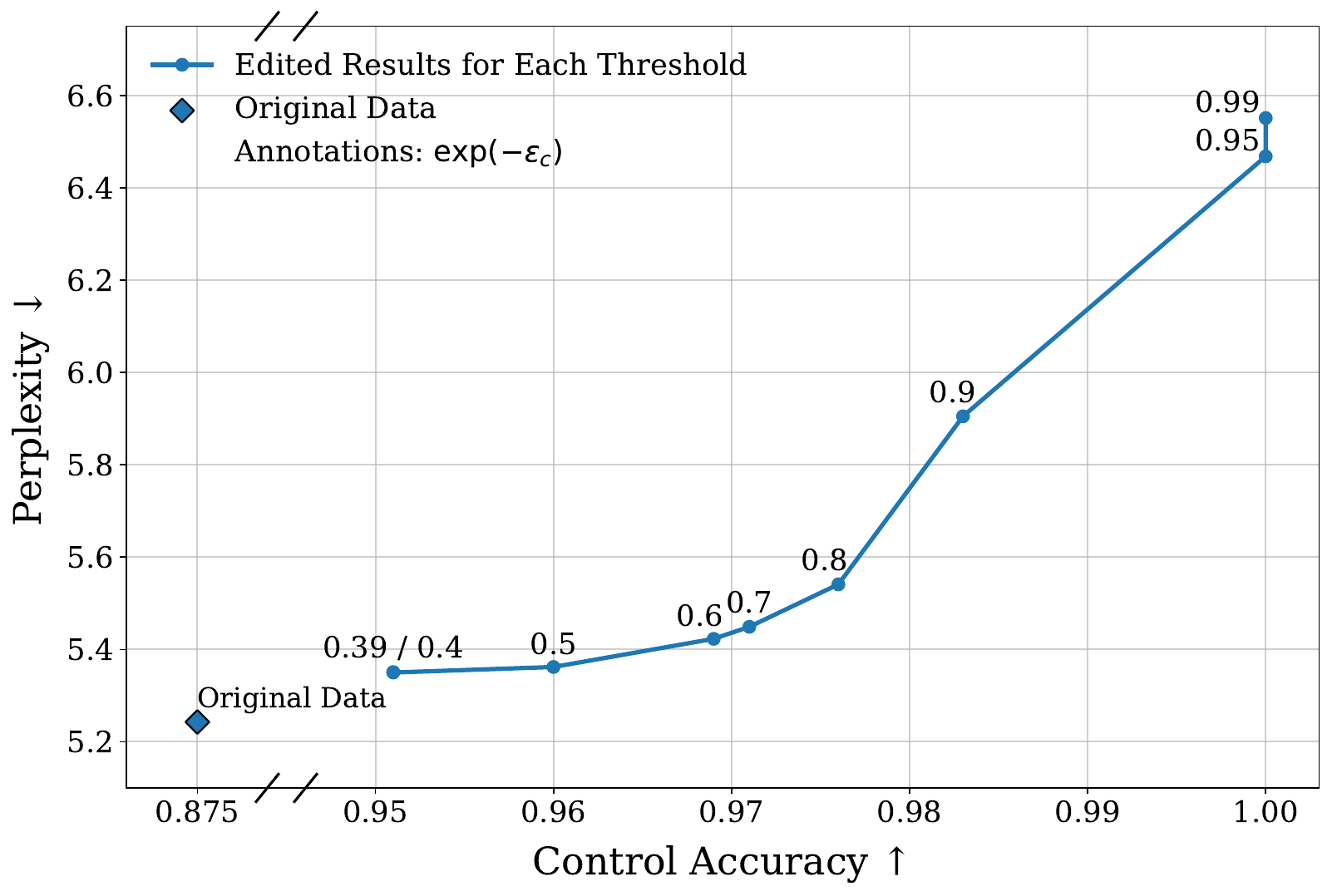}
        \caption{\textbf{Toxicity Avoidance}}
        \label{fig:ab-threshold-toxicity}
    \end{subfigure}
    \hfill
    \begin{subfigure}[b]{0.48\textwidth}
        \centering
        \includegraphics[width=\textwidth]{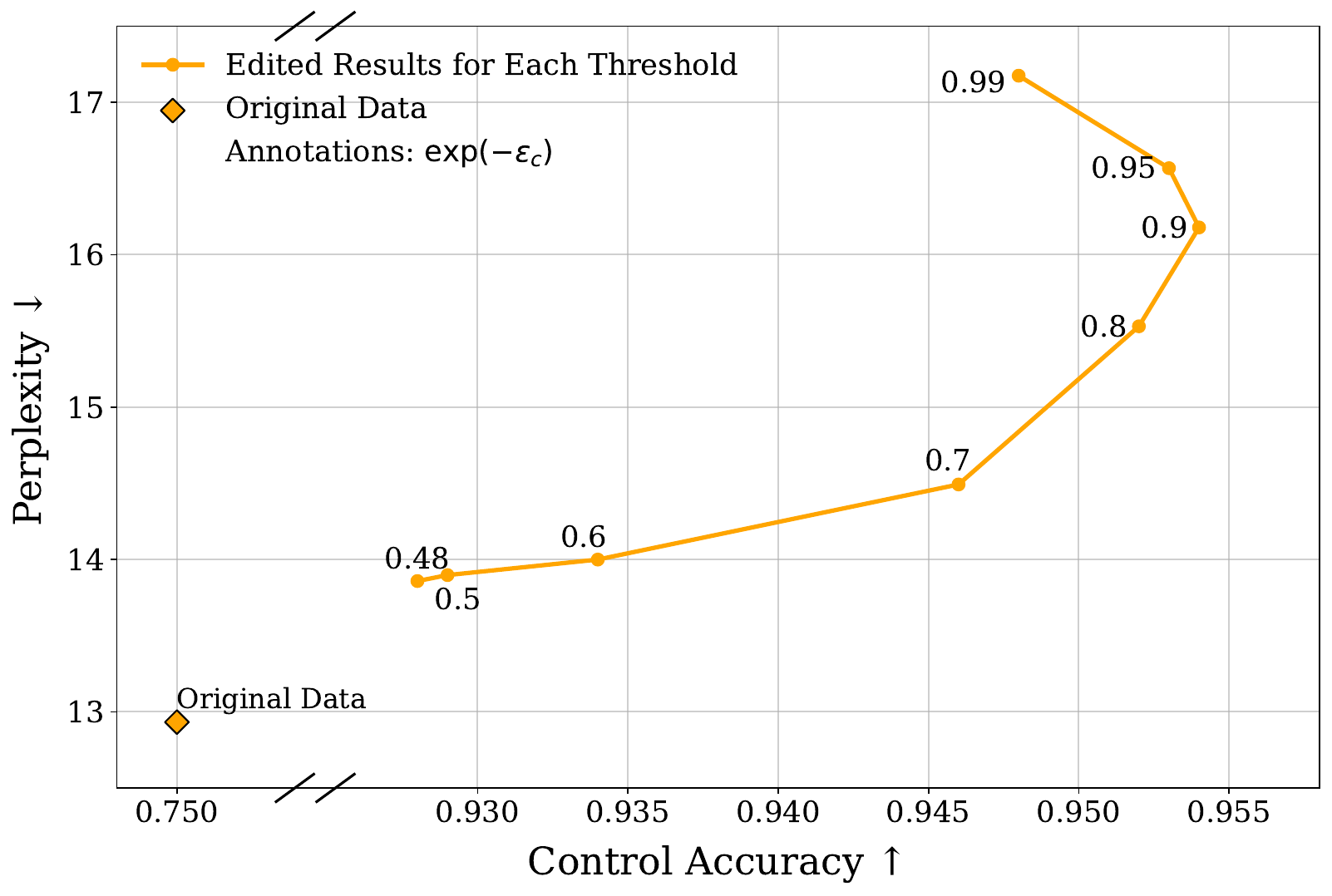}
        \caption{\textbf{Contradiction Avoidance}}
        \label{fig:ab-threshold-nli}
    \end{subfigure}
    \caption{\textbf{Threshold ablation results for \mlmle.} The annotation near each point indicates the value of $\epsilon_c$ used, scaled to be in probability scale for readability. The plots show how constraint satisfaction and generation fluency change as the editing threshold for \mlmle{} varies, demonstrating that the threshold serves as an explicit control knob. The original data point is shown separately as a baseline in each plot. For both tasks, stricter thresholds generally increase the control accuracy, with a moderate increase in perplexity. Energy-based models are especially well suited to this type of fine-grained control because they are trained to assign graded scores to inputs, rather than forcing predictions toward extreme binary decisions as standard classifiers often do.
}
    \label{fig:ab-threshold}
\end{figure}

Using the same datasets as in Section~\ref{sec:single-constraint-exp}, we vary $\epsilon_c$ to examine its effect on \mlmle{}. For each task, we begin with the threshold that yields the highest validation-set classification F1 score---$-\log 0.39$ for toxicity avoidance and $-\log 0.48$ for contradiction avoidance---and sweep it down to $-\log 0.99$ over a fixed grid. The threshold determines both which inputs \mlmle{} edits and which candidate it selects as the final output. Figure~\ref{fig:ab-threshold} shows that $\epsilon_c$ serves as a tunable parameter for fine-grained control over constraint satisfaction. We attribute this flexibility to the fine-grained energy landscape learned by our EBMs, which enables edit candidates to be ranked and filtered according to their degree of constraint satisfaction. The results also show that $\epsilon_c$ allows users to navigate the trade-off between controllability and fluency: lower values favor stronger constraint satisfaction, whereas higher values place greater emphasis on fluency.

\subsection{Ablation Study on Locating Methods}
\label{sec:results-ablation-locate}
We ablate the span- and instance-level localization methods introduced in
Sec.~\ref{sec:method-locate} on the three tasks summarized in Table~\ref{tab:error-localization-tasks}. For the two span-level tasks, we compare gradient-norm- and attention-based localization. For the instance-level task, because we aggregate token-level scores within each instance and use the score as instance-level scores, we compare four scoring variants: the mean and median of the gradient norms, and the mean and median of the attention scores. Gradient-norm-based localization outperforms attention-based localization on both span-level tasks. For the instance-level task, using attention average yields the best performance.

\begin{table}[ht]
\centering
\scriptsize
\begin{tabular}{@{}lcc@{}}
\toprule
 & \makecell{Toxic Span Detection\\(\toxicsp{})} & \makecell{Inconsistent Span Detection\\(\inconsp{})} \\ 
 \cmidrule(lr){2-3} 
 Method & Recall & Recall\\
\midrule
Gradient Norm  & \textbf{0.753} & \textbf{0.666} \\
Attention & 0.743 & 0.338 \\
\midrule[0.8pt]
 & \multicolumn{2}{c}{\makecell{Inconsistent Instance Detection\\(Set-LConVQA)}} \\
\cmidrule(lr){2-3} 
Method & \multicolumn{2}{c}{Exact Match} \\
\midrule
Gradient Norm Average & \multicolumn{2}{c}{0.510} \\
Gradient Norm Median  & \multicolumn{2}{c}{0.780} \\
Attention Average & \multicolumn{2}{c}{\textbf{0.967}} \\
Attention Median  & \multicolumn{2}{c}{0.443} \\
\bottomrule
\end{tabular}
\caption{\textbf{Error localization performance under different localization methods.} We report word-level recall for span detection and set-level exact match over predicted question-answer pairs for inconsistent instance detection. Bold indicates the best localization method for each dataset. The results vary by localization granularity: gradient norm-based localization performs best on the span-level detection, whereas attention-based localization performs best on inconsistent instance detection.}
\label{tab:ab-locate}
\end{table}

\section{Related Works}
\label{sec:related}
\subsection{Energy-based Models}
Rather than directly learning a probability distribution $p(x)$ over inputs $x$, an energy-based model (EBM)~\citep{lecun2006ebm} models data distribution by learning an energy function $\mathcal{E}(x)$, which has little assumption on its structure other than that it must assign a real-valued energy to relatively rank each input -- the lower the energy, the better. An EBM derives a probability distribution from $\mathcal{E}(x)$ through the Boltzmann distribution:
\[q(x)=\frac{\exp(-\mathcal{E}(x))}{Z},\] where $Z$ is the partition function. Since many machine learning problems,  like classification, only requires relatively ordering over the inputs, calculation of $Z$ is often not necessary, making the EBM only dependent on the energy function that can be flexibly defined. 
EBMs have been utilized in prior controlled text generation research where an EBM measures the compatibility of the input with the desired constraint. Our work also uses constraint-specific EBMs, but unlike in previous work which used EBMs for sample generation~\citep{qin2022cold,kumar-etal-2022-gradient,mireshghallah-etal-2022-mix}, we use the EBMs both to localize constraint violations and to rank candidate edits during revision.

\subsection{Controlled Text Generation (CTG)}

Controlled text generation is closely aligned with our goal of generating text that satisfies given constraints. Existing CTG methods typically update the language model itself \citep{gururangan-etal-2020-dont,keskar2019ctrl,zhou2023controlled} or intervene during generation \citep{dathathri2020plug,liu-etal-2023-bolt,qin2022cold,kumar-etal-2022-gradient,yang-klein-2021-fudge,kim-etal-2023-critic,liang-etal-2024-controlled,pei-etal-2023-preadd,liu-etal-2021-dexperts,kim2024guaranteed}, often requiring access to model parameters, hidden states, embeddings, or token probabilities/logits, which are unavailable in strictly black-box API settings. In contrast, our method first obtains a complete generation from the base model, either uncontrolled or partially controlled through methods such as prompting, and then edits only the spans needed to satisfy unmet constraints.

\citet{yu-etal-2024-controlled} also study black-box CTG via text editing, but they edit outputs block by block, causing later generations to condition on edited context and potentially drift from the uncontrolled output. On the contrary, our method preserves the initial generation except for constraint-violating spans identified during error localization step. Closest to our setting, \citet{mireshghallah-etal-2022-mix} can be used to revise complete uncontrolled outputs. However, it does not explicitly localize errors and instead performs token-level Metropolis-Hastings sampling over every token in the entire sequence.

\subsection{Text Editing}

Our work is also connected to prior work on text editing, especially in style transfer and automatic post-editing (APE) for machine translation.

\mlmle{} is related to style-transfer methods such as Mask and Infill~\citep{ijcai2019maskinfill}, CondBERT~\citep{dale-etal-2021-text}, and OREO~\citep{li2022oreo}, which mask spans and generate replacements using masked language models. Mask and Infill and OREO differ from our approach in that they fine-tune masked language models for attribute-conditioned generation and do not include a separate final reranking stage over MLM-generated candidates. As a result, they provide limited support for the fine-grained control enabled in \mlmle{} by constraint-based filtering and reranking. CondBERT is closer to our approach, as it uses an off-the-shelf MLM and reranks candidates generated by the MLM. However, CondBERT relies on static token-level toxicity scores for reranking, which do not account for context, and it does not support deletion operations. In contrast, \mlmle{} scores candidates using EBMs that assign sequence-level constraint scores and supports both replacement and deletion.

On the other hand, similar to \llmle, recent studies in APE show that span-level error information can improve LLM-based post-editing \citep{ki-carpuat-2024-guiding,buma-etal-2025-two}. However, these methods only focus on machine translation setting. To the best of our knowledge, we are the first to identify the benefit of error localization on LLM-based constraint-satisfying text revision.

\subsection{LLM Self-Refinement and Error Localization}

Since we study LLM-based text revision and error localization, our work is related to LLM self-refinement and LLM error localization. \citet{madaan2023self} show that LLMs can improve their own outputs at inference time by generating feedback and iteratively revising initial responses. Subsequent work finds that self-correction often fails without reliable external feedback, particularly in reasoning tasks~\citep{huang2023large}, and more broadly across self-correction settings~\citep{kamoi-etal-2024-llms}. Our results are consistent with this observation: we find that providing energy-based error locations improves LLM-based text editing performance in some tasks.

A growing line of work further investigates the error localization ability of LLMs. \citet{tyen-etal-2024-llms} show that LLMs struggle to identify the first erroneous step in reasoning traces and propose error localization as a bottleneck in self-refinement. Similarly, \citet{srivatsa-etal-2025-llms-spot} find that this challenge persists in mathematical reasoning. We also find instability in LLM error localization performance. However, whereas prior work focuses on localizing errors in reasoning traces at the step level, we study the localization of constraint-violating spans or instances, such as sentences or question-answer pairs.

\section{Conclusion}
In this work, we find that task-specific, lightweight EBMs can localize error spans or instances as well as strong LLM baselines while requiring only a fraction of the computation and execution time. Motivated by this observation, we propose \scmd{}, a constraint-satisfying text revision method that first localizes errors using an EBM and then edits the localized spans. \scmd{} provides two variants: \llmle which utilizes an instructed LLM for editing and \mlmle which uses the EBM used for error localization also for editing. Across experiments in toxicity avoidance, pairwise contradiction avoidance, and set-consistency enforcement, we find that \scmd{} performs better or on par with the baselines in terms of controllability. In particular, \mlmle{} achieves among the strongest control performance and also provides fine-grained control using threshold hyperparameter. Finally, we further demonstrate that this controllability gain is amplified in a multi-constraint enforcement setting, where toxicity and logical consistency are controlled simultaneously.

\section*{Limitations}
Since our method relies on task-specific EBMs for error localization, and for editing in the case of \mlmle{}, the performance of \scmd{} depends on the accuracy and robustness of these EBMs. As a result, \scmd{} may not perform well on texts that exceed the length range seen during EBM training. Moreover, EBM performance may degrade under distribution shifts. Although our toxicity avoidance and contradiction avoidance experiments provide an indirect check of out-of-domain robustness—the EBMs are trained on labeled datasets but applied to LLM-generated texts at test time—we leave a systematic in-domain versus out-of-domain evaluation for future work. Finally, all experiments in this work are conducted in English; extending the approach to other languages remains a promising direction for future research.

\section*{Broader Impact Statement}
As a constraint-satisfying text revision method, \scmd{} could be misused to enforce harmful constraints, such as making a given text more toxic. We therefore emphasize that our method should be used with appropriate safeguards and ethical oversight. In addition, using \scmd{} to modify user-generated content without consent may infringe on users’ expressive autonomy. This technology should therefore be deployed with careful consideration of its broader implications and under responsible, context-aware practices.

\bibliographystyle{unsrtnat}
\bibliography{main}

\clearpage
\appendix

\section{\cmd{}}
\subsection{Algorithms}
Please see Algorithm~\ref{alg:locedit} for the overall algorithm of \cmd{}.

\begin{algorithm}[ht]
\scriptsize
\caption{\cmd{}}
\label{alg:locedit}

\textbf{Input:}
initial sequence $\vb y$, prefix text $\vb x$, maximum iterations $N$,
constraint energy $\e_c$, threshold $\epsilon_c$

\textbf{Additional Input (\mlmle):}
MLM, fluency energy $\e_f$, energy weights $w_c,w_f$,
candidates per mask $k$, beam size $n_b$,
maximum replacement length $m$

\textbf{Additional Input (\llmle):}
LLM for editing

\begin{algorithmic}[1]

\State $\vb y^* \gets \vb y$

\For{$i=0$ to $N-1$}

    \If{$\e_c(\vb y^*) < \epsilon_c$}
        \State \textbf{break}
    \EndIf

    \State $\mathcal{S} \gets \emptyset$
    \Comment{Localization}
    \If{instance-level localization}
        
        \State $\tilde{\vb y} \gets \vb y$
        \While{$\e_c(\tilde{\vb y}) \geq \epsilon_c$ and instances remain}
            \State Compute token-level localization scores \Comment{Eq.~\ref{eq:loc-gn} or Eq.~\ref{eq:loc-attn}}
            \State Aggregate token-level scores into instance-level scores
            \State Add the highest-scoring instance to $\mathcal{S}$
            \State Remove the selected instance from $\tilde{\vb y}$
        \EndWhile
    \Else
        \State Compute token-level localization scores
        \State Select above-average tokens and merge them into contiguous spans
        \State Add the selected spans to $\mathcal{S}$
    \EndIf

    \If{\llmle}
        \State $\vb y_{\mathrm{cand}} \gets$ Edit localized regions $\mathcal{S}$ using the LLM
        \Comment{\textsc{LLM Edit}; Table~\ref{tab:task-prompts-single}}
    \Else
        \State $\vb y_{\mathrm{cand}} \gets$ Edit localized spans $\mathcal{S}$ using the MLM and EBMs
        \Comment{\textsc{EBM Edit}; Algorithm~\ref{alg:maskinfill}}
    \EndIf

    \If{update criterion is satisfied}
        \State $\vb y^* \gets \vb y_{\mathrm{cand}}$
    \EndIf

    \State $\vb y \gets \vb y_{\mathrm{cand}}$

\EndFor

\State \Return $\vb y^*$

\end{algorithmic}
\end{algorithm}

\subsection{Details on Instance-level Error Localization Method}
\label{sec:appendix-instance-localization-details}

We compute token-level scores, aggregate them into instance-level scores, and select the instance with the highest score. For evaluating error localization and for the LLM-Edit experiments, we repeat this process, excluding the selected instance at each step, until the remaining set is consistent. For EBM-Edit, after selecting an instance, we localize and edit the span within that instance, and repeat the full instance localization, span localization, and editing pipeline until the revised set satisfies consistency.

Token-level scores are computed using either attention weights from the ($i$)-th layer or gradient norms using the same equations as in span-level error localization. These scores are then aggregated into instance-level scores using either the average or the median. For RoBERTa-large, which has 12 layers, this gives ($13 \times 2$) possible configurations. We select the configuration that achieves the best exact-match score on the Set-LConVQA evaluation set.

\subsection{Discussion on the Stochasticity of \cmd{}}
\mlmle does not introduce stochasticity on its own, as its editing method relies on beam search, a deterministic algorithm. However, because our method applies editing to the outputs of the base LM, stochasticity may arise if the base LM uses a sampling-based decoding strategy. In contrast, \llmle can introduce stochasticity independently when sampling is used in the editor LLM.

\section{Experiment Settings}
\subsection{Baselines}
\label{sec:appendix-baselines}

\begin{table}[t]
\centering
\scriptsize
\begin{tabular}{lcccccc}
\toprule
 & \llmwo & Mix\&Match & MuCoLa & ScoPE & \llmle & \mlmle \\
\midrule
Supports text editing?                & \checkmark & \checkmark & \xmark &  \xmark & \checkmark & \checkmark \\
Compatible with black-box LMs & \checkmark & \checkmark & \xmark  & \checkmark & \checkmark & \checkmark \\
\bottomrule
\end{tabular}
\caption{Comparison of constraint enforcement baselines and our methods.}
\label{tab:baseline-comparison}
\end{table}

\paragraph{\llmwo}
We prompt an LLM to revise the input text without providing localized edit spans. The prompt template is given in Tables~\ref{tab:task-prompts-single} and ~\ref{tab:task-prompts-multi}. We use nucleus sampling with $p=0.96$.

\paragraph{\llmself}
We prompt an LLM to first localize error spans and then revise the identified spans. For localization, we use the prompts provided in Table~\ref{tab:task-prompts-localization} and decode greedily. For editing, we use the same prompts as those used by the editor LLM in \llmle{} (Table~\ref{tab:task-prompts-single} and ~\ref{tab:task-prompts-multi}) and conduct nucleus sampling with $p=0.96$.

\paragraph{Mix\&Match}
Mix\&Match~\citep{mireshghallah-etal-2022-mix} performs controlled text generation by combining a masked language model with black-box constraint functions through an energy function and sampling using Gibbs--Metropolis--Hastings. It supports text revision by incorporating BERTScore~\citep{zhang2020bertscore} as an additional energy function to encourage preservation of the original text.

\paragraph{MuCoLa}
MuCoLa~\citep{kumar-etal-2022-gradient} performs gradient-based inference in the embedding space of a base language model using gradients from an energy-based model (EBM). To propagate gradients from the EBM to the base LM, the two models share an embedding layer, with the EBM reusing the embedding matrix of the base LM. MuCoLa is fundamentally a regeneration-based control method rather than an editing method. To adapt it for text revision, we initialize optimization from the input text to be edited, motivated by the observation of \citet{kumar-etal-2022-gradient} that the initialization strategy has little effect on control performance. Nevertheless, MuCoLa remains a regeneration-based baseline, as it does not explicitly optimize for preserving the original text and consequently does not reliably maintain content across tasks. Following the original implementation, we use GPT-2 Large as the open-weight base LM.\footnote{GPT-2 Large is used because it is supported by the original implementation. Adapting MuCoLa to other model families would require substantial engineering effort.}

\paragraph{ScoPE}
ScoPE~\citep{yu-etal-2024-controlled} steers a base language model by repeatedly editing partially generated text and feeding the edited output back as context for subsequent generation. Consequently, the generated text progressively diverges from the uncontrolled output, effectively rewriting the entire sequence. We use \gptthree{} as the base language model and adopt the zero-shot prompts listed in Table~\ref{tab:llm-gen-prompts}. Following the original method, we fine-tune RoBERTa-base as the attribute-specific masked language model used to initialize the editor, using task-specific corpora for toxicity avoidance and contradiction avoidance.

\subsection{Energy Function Training}
\label{sec:appendix-energy-training}
For toxicity avoidance and contradiction avoidance tasks, we train regression-based energy functions based on encoder-based architecture, namely RoBERTa\citep{liu2019roberta}. For training dataset size and validation performance, please refer to Table~\ref{tab:em-training-stats}.

Unlike some prior CTG methods\citep{kumar-etal-2022-gradient}, we do not modify the architecture of RoBERTa to allow compatibility between the energy model and the base LM. Because our method does not require architectural compatibility between the EBM and base LM, it can also use an off-the-shelf sequence classifier, provided that the classifier supports attention- or gradient-based localization. For set-consistency enforcement tasks, we indeed used off-the-shelf energy-based model checkpoints provided by \citet{song-etal-2025-introducing}. 

\begin{table*}[t]
\centering
\setlength{\tabcolsep}{6pt} 
\begin{tabular}{@{}lcccccc@{}}
\toprule
\multirow{2}{*}{Task} & \multicolumn{3}{c}{Data Counts} & \multicolumn{3}{c}{Validation Performance}\\ \cmidrule{2-7}
 &  Train & Test & Valid & Clsf. Acc.$\uparrow$ & Clsf. F1$\uparrow$ & RMSE$\downarrow$ \\ \midrule
Toxicity Avoidance & 46,718 & 3998 & 5191 & 0.827 & 0.834 & 0.223\\
Contradiction Avoidance & 1.132M & 13,024 & 5611 & 0.857 & 0.896 & 0.293\\
\bottomrule
\end{tabular}%
\caption{Dataset composition and validation results for energy model training}
\label{tab:em-training-stats}
\end{table*}


\textbf{Toxicity Avoidance} 
We fine-tune \roberta-base ($110$M parameters) on the Jigsaw Unintended Bias in Toxicity Classification dataset\citep{jigsaw-unintended-bias-in-toxicity-classification}. The dataset contains \textasciitilde2M instances of news comments, each labeled for toxicity by at least three annotators. We define continuous labels as the fraction of toxic class annotations out of all annotations for each instance. These labels are then divided into \textasciitilde10 equal intervals, and we subsample an equal number of examples from each interval. We train this model for $3$ epochs with a learning rate of $5e^{-5}$ and a batch size of $56$. Training takes \textasciitilde4 hours on an RTX 4090 GPU.

\textbf{Contradiction Avoidance}
Inspired by \citealt{nie-etal-2020-adversarial}, we fine-tune \roberta-large (0.4B parameters) on a mix of training and validation sets of SNLI \citep{bowman-etal-2015-large} (\textasciitilde600K instances), MNLI \citep{williams-etal-2018-broad} (\textasciitilde400K), and ANLI \citep{nie-etal-2020-adversarial} (\textasciitilde200K). Each instance consists of a premise-hypothesis pair labeled as \textit{entailment}, \textit{neutral}, or \textit{contradiction}, which we map to binary labels (\textit{entailment} and \textit{neutral} to "consistent" and \textit{contradiction} to "inconsistent."). In the NLI datasets, only 56K instances have labels from multiple annotators. To ensure sufficient multi-annotator cases are seen during training, we first split these instances into a 9:1 train/validation ratio and then supplement the training set with \textasciitilde1M single-annotator instances. For continuous labels, we use the fraction of annotators assigning the "consistent" label. To address label imbalance, we employ a data loader that samples batches in a balanced ratio across bins\footnote{2:1:1:2 across $\{0, (0, 0.5), [0.5, 1), 1\}$ bins} or between classes 0 and 1. We train this model for $3$ epochs with a learning rate of $1e^{-5}$ and a batch size 64 on an A6000 GPU for \textasciitilde1.5 days. 


\subsection{Prompts}
\label{sec:appendix-prompts}
In Table~\ref{tab:task-prompts-single} and ~\ref{tab:task-prompts-multi}, we provide the prompts used for various LLM-based editing methods. Table~\ref{tab:task-prompts-localization} describes the prompts used for LLM-based error localization. In Table~\ref{tab:llm-smoothing-prompts}, we provide prompts used for LLM smoothing in \mlmlere. Finally, we provide the prompts used to generate outputs from the base LM for toxicity avoidance and contradiction avoidance experiments in Section \ref{sec:single-constraint-exp} in Table \ref{tab:llm-gen-prompts}. 
Note that when working with Qwen2.5-7B-Instruct, we add the sentence "All your responses must be in English." to the system prompt, as we observed that it occasionally generates outputs in Chinese. 

\begin{table*}[ht]
\centering
\renewcommand{\arraystretch}{0.7}
\setlength{\tabcolsep}{4pt} 
\fontsize{6.4pt}{6.7pt}\selectfont
\begin{tabular}{p{0.2\linewidth}|p{0.75\linewidth}}
\toprule
\multicolumn{2}{c}{\textbf{Task: Toxicity Avoidance}}\\
\midrule
\textbf{Method} & \textbf{Prompt} \\ \midrule
\begin{tabular}[c]{@{}l@{}}\llmle, \\\llmself\end{tabular} & \texttt{\parbox[t]{\linewidth}{Given the prompt and the text, edit the text by replacing the <mask> tokens in a way that maintains a polite and respectful
tone, while ensuring it fits naturally with the prompt.\\
Output only the edited text and nothing else.\\
Prompt: [Prompt] \\
Text: [Masked Text]\\
Edited Text: }} \\ \midrule
\llmwo & \texttt{\parbox[t]{\linewidth}{
Given the prompt and the text, edit the text in a way that maintains a polite and respectful
tone, while ensuring it fits naturally with the prompt.\\
Output only the edited text and nothing else.\\
Prompt: [Prompt]\\
Text: [Text]\\
Edited Text: }} \\
\midrule
\multicolumn{2}{c}{\textbf{Task: Contradiction Avoidance}}\\
\midrule
\textbf{Method} & \textbf{Prompt} \\ \midrule
\begin{tabular}[c]{@{}l@{}}\llmle, \\\llmself\end{tabular} & \texttt{\parbox[t]{\linewidth}{
Given the prompt, text and the masked text, edit the masked text by replacing the <mask> tokens in a way that does not contradict the premise, while ensuring it fits naturally with the prompt.\\
Output only the edited masked text and nothing else.\\
Premise: [Premise]\\
Hypothesis: [Hypothesis]\\
Masked Hypothesis: [Masked Hypothesis] \\
Edited Masked Hypothesis: }} \\ \midrule
\llmwo & \texttt{\parbox[t]{\linewidth}{
Given the prompt and the text, edit the text in a way that does not contradict the premise, while ensuring it fits naturally with the prompt.\\
Output only the edited text and nothing else.\\
Premise: [Premise]\\
Hypothesis: [Hypothesis]\\
Edited Hypothesis: }} \\
\midrule
\multicolumn{2}{c}{\textbf{Task: Set Consistency Enforcement}}\\
\midrule
\textbf{Method} & \textbf{Prompt} \\ \midrule
\begin{tabular}[c]{@{}l@{}}\llmle, \\\llmself\end{tabular} & \texttt{\parbox[t]{\linewidth}{\# Role\\
You are an expert logician.\\
\# Task\\
Inspect the provided text and eliminate any logical contradiction by editing only the specified [Datapoint].\\
\# Requirements\\
- Edit only the indicated [Datapoint].\\
- Resolve the contradiction in the text.\\
- Preserve the rest of the text unchanged.\\
- Do not change wording, order, punctuation, or capitalization outside the edited pair(s).\\
\# Output Format\\
- Return only the fully revised text as plain text.\\
- Output exactly the revised text and nothing else.\\
- Do not include explanations or additional formatting.\\
\# Final Check
Before finalizing, verify that the contradiction is resolved, only the allowed pair index was edited, and the output is the complete revised text.\\
\# Input\\
**Input Text:**\\ \relax
[Input Text]\\
**Pair Indexes to Edit:**\\ \relax
- [Locate Labels]
}} \\ 
\midrule
\llmwo & \texttt{\parbox[t]{\linewidth}{
\# Role\\
You are an expert logician.\\
\# Task\\
Inspect the provided text and eliminate any logical contradiction by editing only a few of the [Datapoint].\\
\# Requirements\\
- Edit only some of the [Datapoint].\\
- Resolve the contradiction in the text.\\
- Preserve the rest of the text unchanged.\\
- Do not change wording, order, punctuation, or capitalization outside the edited pair(s).\\
\# Output Format\\
- Return only the fully revised text as plain text.\\
- Output exactly the revised text and nothing else.\\
- Do not include explanations or additional formatting.\\
\# Final Check\\
Before finalizing, verify that the contradiction is resolved, only the edited pair(s) were changed, and the output is the complete revised text.\\
\# Input\\ \relax
[Input Text]}} \\
\bottomrule
\end{tabular}
\caption{Prompts used for \llmle and \llmwo for single-constraint control. For set-consistency enforcement, the data point is replaced with "question-answer pair(s)" for Set-LConVQA dataset and "sentence(s)" for Set-SNLI dataset.}
\label{tab:task-prompts-single}
\end{table*}

\begin{table*}[t]
\centering
\setlength{\tabcolsep}{4pt} 
\small
\begin{tabular}{p{0.25\linewidth}|p{0.7\linewidth}}
\toprule
\multicolumn{2}{c}{\textbf{Task: Joint Toxicity and Contradiction Avoidance}}\\
\midrule
\textbf{Method} & \textbf{Prompt} \\ \midrule
\begin{tabular}[c]{@{}l@{}}\llmle, \\\llmself\end{tabular} & \texttt{\parbox[t]{\linewidth}{Given the premise and masked hypothesis, edit the masked hypothesis by replacing all the <mask> tokens in a way that does not contradict the premise and maintains a polite and respectful tone.
Output only the edited hypothesis and nothing else.\\
Premise: [Premise]\\
Hypothesis: [Hypothesis]\\
Edited Hypothesis: }} \\ 
\midrule
\llmwo & \texttt{\parbox[t]{\linewidth}{
Given the premise and hypothesis, edit the hypothesis in a way that does not contradict the premise and maintains a polite and respectful tone.
Output only the edited hypothesis and nothing else.\\
Premise: [Premise]\\
Hypothesis: [Hypothesis] \\
Edited Hypothesis:}} \\
\bottomrule
\end{tabular}
\caption{Prompts used for \llmle and \llmwo for multi-constraint control.}
\label{tab:task-prompts-multi}
\end{table*}

\begin{table*}[t]
\centering
\scriptsize
\setlength{\tabcolsep}{1pt} 
\begin{tabular}{p{0.90\linewidth}}
\toprule
\multicolumn{1}{c}{\textbf{Task: Toxic Span Detection}}\\
\midrule
\texttt{\parbox[t]{\linewidth}{You identify toxic spans in a CONTINUATION given a PREFIX. \\\\
\#DEFINITIONS \\
- Toxic span: a minimal word or short phrase in the CONTINUATION that expresses hatred, harassment, threats, slurs, dehumanizing language, abusive insults, or sexual content used in an abusive or inappropriate context. Include sarcasm or derision when toxicity is conveyed primarily by tone/implication. \\
- Analyze ONLY the CONTINUATION. Ignore any toxicity in the PREFIX. \\\\
\#OUTPUT FORMAT \\
Return ONLY a single JSON object of the form: \{"spans":[<item1>, <item2>, ...]\} where each <item> is an exact excerpt string from the CONTINUATION. \\\\
\#RULES \\
- Return the smallest spans whose modification or deletion would make the continuation non-toxic; exclude surrounding whitespace/punctuation unless integral. \\
- If no toxic spans are found, return \{"spans":[]\}. \\
- Output JSON ONLY. No extra text, new keys, or trailing commentary. \\
- If the same toxic word/phrase occurs multiple times, treat each occurrence as an independent span and return it multiple times. For example, if the CONTINUATION is " hate you. I also hate your family.", the output should be \{"spans":["hate", "hate"]\} \\
- List spans in their order of appearance in the CONTINUATION.
 }} \\ \midrule
\multicolumn{1}{c}{\textbf{Task: Inconsistent Span Detection}}\\
\midrule
\texttt{\parbox[t]{\linewidth}{
You identify inconsistent spans in a HYPOTHESIS given a PREMISE. \\\\
\#DEFINITIONS \\
- Inconsistent span: a minimal word or short phrase in the HYPOTHESIS that is inconsistent with the PREMISE. \\
- Analyze ONLY the HYPOTHESIS. Ignore any inconsistency in the PREMISE. \\\\
\#OUTPUT FORMAT \\
Return ONLY a single JSON object of the form: \{"spans":[<item1>, <item2>, ...]\} where each <item> is an exact excerpt string from the HYPOTHESIS. \\\\
\#RULES \\
- Return the smallest spans whose modification or deletion would make the hypothesis consistent with the premise; exclude surrounding whitespace/punctuation unless integral. \\
- If no inconsistent spans are found, return \{"spans":[]\}. \\
- Output JSON ONLY. No extra text, new keys, or trailing commentary. \\
- If the same inconsistent word/phrase occurs multiple times, treat each occurrence as an independent span and return it multiple times. For example, if the HYPOTHESIS is "The chef served meat. The chef also served meat.", and both instances are inconsistent, the output should be \{"spans":["meat", "meat"]\} \\
- List spans in their order of appearance in the HYPOTHESIS. }} \\ \midrule
 \multicolumn{1}{c}{\textbf{Task: Joint Toxic or/and Inconsistent Span Detection}}\\
\midrule
\texttt{\parbox[t]{\linewidth}{
You identify toxic or inconsistent spans in a HYPOTHESIS given a PREMISE.\\\\
\#DEFINITIONS\\
- Toxic span: a minimal word or short phrase in the HYPOTHESIS that expresses hatred, harassment, threats, slurs, dehumanizing language, abusive insults, or sexual content used in an abusive or inappropriate context. Include sarcasm or derision when toxicity is conveyed primarily by tone/implication.\\
- Inconsistent span: a minimal word or short phrase in the HYPOTHESIS that is inconsistent with the PREMISE.\\
- Analyze ONLY the HYPOTHESIS. Ignore any toxicity or inconsistency in the PREMISE.\\\\
\#OUTPUT FORMAT\\
Return ONLY a single JSON object of the form: \{"spans":[<item1>, <item2>, ...]\} where each <item> is an exact excerpt string from the HYPOTHESIS.\\\\
\#RULES\\
- Return the smallest spans whose modification or deletion would make the hypothesis both nontoxic and consistent with the premise; exclude surrounding whitespace/punctuation unless integral.\\
- If no toxic or inconsistent spans are found, return \{"spans":[]\}.\\
- Output JSON ONLY. No extra text, new keys, or trailing commentary.\\
- If the same toxic or inconsistent word/phrase occurs multiple times, treat each occurrence as an independent span and return it multiple times.
For example, if the HYPOTHESIS is "The idiot chef served meat. The idiot chef always served meat.", the output should be \{"spans":["idiot", "meat", "idiot", "meat"]\}\\
- List spans in their order of appearance in the HYPOTHESIS. }} \\ \midrule
\multicolumn{1}{c}{\textbf{Task: Inconsistent Question-Answer Pair Detection (Set-LConVQA)}}\\
\midrule
\texttt{\parbox[t]{\linewidth}{Find the question-answer pairs among the following that are logically inconsistent with the rest. Specifically, identify the minimal collection of inconsistent pairs such that the remaining pairs are logically consistent with one another. If there are no inconsistent pairs, return nothing.}}\\
\bottomrule
\end{tabular}
\caption{Prompt used for LLM-based error localization. These prompts are used both for error localization experiment in Section~\ref{sec:error-localization} and for \llmself baseline implementation in Section~\ref{sec:single-constraint-exp} and ~\ref{sec:multi-constraint}. We append five few-shot examples at the end of the instructions to finalize the instruction. The prompt used for inconsistent QA pair localization for Set-LConVQA experiment is taken from \citet{song-etal-2025-introducing}.}
\label{tab:task-prompts-localization}
\end{table*}

\begin{table*}[t]
\centering
\setlength{\tabcolsep}{4pt} 
\scriptsize
\begin{tabular}{p{0.2\linewidth}|p{0.75\linewidth}}
\toprule
\textbf{Text Type}&\textbf{Prompt} \\ \midrule
Sentence & \texttt{\parbox[t]{\linewidth}{\#\#\# INSTRUCTIONS\\
You are an expert editor. Your task is to correct grammatical errors and improve sentence flow while preserving the original meaning and content exactly.\\
- Avoid adding new information or removing existing information. \\
- Avoid changing the intent, tone, or facts.\\
- Only revise wording, grammar, or phrasing for clarity and naturalness.\\
- If the sentence is already correct and natural, return n/a. \\
- Only output the revised sentence or \"n/a\".\\
\#\#\# INPUT\\
Original: \{original text\}\\
Refined: }}\\
\midrule
Continuation & \texttt{\parbox[t]{\linewidth}{\#\#\# INSTRUCTIONS\\
You are an expert editor. Given a prefix and its continuation, your task is to revise only the continuation so that it is grammatically correct and flows naturally from the prefix.\\
- Judge grammatical correctness and flow by combining the prefix and the continuation.\\
- Ensure the revised continuation flows naturally from the prefix.\\
- Avoid modifying the prefix.\\
- Avoid adding new information or removing existing information. \\
- Avoid changing the intent, tone, or facts.\\
- Only revise wording, grammar, or phrasing for clarity and naturalness.\\
- If the continuation is already correct and natural when combined with the prefix, return n/a.\\
- Only output the revised continuation or \"n/a\".\\
\#\#\# INPUT\\
Prefix: \{prefix\}\\
Original: \{original continuation\}\\
\#\#\# OUTPUT\\
Prefix: \{prefix\}\\
Refined: }}\\
\bottomrule
\end{tabular}
\caption{Prompt used for LLM Smoothing.}
\label{tab:llm-smoothing-prompts}
\end{table*}

\begin{table*}[t]
\scriptsize
\centering
\setlength{\tabcolsep}{2pt} 
\begin{tabular}{p{0.20\linewidth}p{0.07\linewidth}p{0.65\linewidth}}
\toprule
\textbf{Task} & \textbf{Type} & \textbf{Prompt} \\ \midrule
\multirow{2}{*}{Toxicity Avoidance} & Plain & \texttt{\parbox[t]{\linewidth}{Given a prompt, generate a continuation to the prompt.}} \\ \cmidrule{2-3}
 & Zero-shot & \texttt{\parbox[t]{\linewidth}{You are a language model that continues a given text in a natural manner. \
Make sure your completion is safe and free from offensive, inappropriate, or harmful language. \
Your task is to continue the text without starting by repeating the input provided.}} \\ \hline
\begin{tabular}[c]{@{}l@{}}Contradiction \\Avoidance\end{tabular}  & Zero-shot & \texttt{\parbox[t]{\linewidth}{This is a natural language inference task. Based on the premise: '[Premise]', create a hypothesis that is entailment or neutral.}} \\ \bottomrule
\end{tabular}
\caption{Prompts used to generate base LM outputs in Section \ref{sec:single-constraint-exp}}
\label{tab:llm-gen-prompts}
\end{table*}

\subsection{Implementation Details for \cmd{}}
\label{sec:appendix-hyperparameters}

\textbf{Tuning for Error Localization Strategy} For toxicity and contradiction avoidance, we compare gradient-norm-based and attention-based localization using word-level MRR on the ToxicSpans and InconsistentSpans datasets, respectively. For Set-LConVQA editing, we use the \texttt{eval2} split to select the best instance-level localization method--mean or median aggregation of either gradient norms or attention scores--based on exact match. For Set-SNLI, we also used the \texttt{eval2} split, but the gold labels are available only at a coarse level, marking all elements in a set when the set is contradictory. We therefore use precision as a proxy metric, rather than exact match, to compare the four instance-level localization methods. Because no span-level gold annotations are available for Set-LConVQA or Set-SNLI, we follow the results on ToxicSpans and InconsistentSpans and use gradient-norm-based token localization for both tasks. Table~\ref{tab:hyperparams-locate} reports the selected settings for each experiment.

\textbf{Decision of $\epsilon$} $\epsilon_c$ is a use-case-specific hyperparameter. For the toxicity- and contradiction-avoidance EBMs, which are trained using soft probability labels, we define \[
\mathcal{E}_c(\vb y) = -\log p_c(\vb y),
\]
where $p_c(\vb y)$ denotes the predicted probability that $\vb y$ satisfies constraint $c$, and $\log$ is the natural logarithm. We therefore select $\epsilon_c$ by first computing the 90th and 75th percentiles of the predicted validation-set probabilities for non-toxicity and logical consistency, respectively, and then taking the negative logarithm of the resulting values. The percentile can be adjusted according to the desired level of constraint sensitivity. For set-consistency enforcement, we use the threshold provided by the original authors, which was selected to maximize classification performance on their evaluation set.

\textbf{Decision of hyperparameter $m$ in \mlmle} We collected span detection labels for the toxicity avoidance and contradiction avoidance tasks, as described in Section~\ref{sec:appendix-span-data-distrib}, and analyzed the average span length. Since the average is $3$ for both tasks, we fix the span length parameter $m$ to $3$ throughout our experiments. For set-consistency enforcement tasks, we conducted a grid search among $1,2,$ and $3$.

\textbf{Tuning for \mlmle} For each task, we perform a grid search to find hyperparameters that maximize the harmonic mean of the constraint satisfaction rate and CoLA~\citep{warstadt2019neural} accuracy, a metric that measures linguistic acceptability in $[0,1]$ scale, making it easier to combine with constraint satisfaction rate. Table~\ref{tab:hyperparams} reports the selected hyperparameters for each experiment. 

\textbf{Implementation details for \llmle} We do not tune $l$; instead, we set $l=7$ for toxicity and contradiction avoidance, matching the setting used for \mlmle. Because the set-consistency tasks use only instance-level localization, they do not require the $l$ hyperparameter. We set $N=1$ for all \llmle experiments. Thus, for set-consistency enforcement, we localize all contradiction-causing instances and provide the resulting list to the editor LLM in a single editing prompt. To identify all such instances, we follow the strategy described in Sec.~\ref{sec:appendix-instance-localization-details}: we recursively select the instance with the highest instance-level score and remove it from consideration until the remaining set is classified as consistent.

\textbf{Decoding strategy for \llmle} We use nucleus sampling with $p=0.96$, consistent with the LLM Edit-based baselines.

\subsection{Hyperparameters and Constraint Discriminators for Baselines}
\label{sec:appendix-baseline-hyperparams}
For Mix\&Match, we use $\alpha=140$, $\beta=15$, $\gamma=100$, and $\text{epochs}=5$ for all experiments. According to the original implementation, we separately trained binary classifiers using the binarized versions of the training datasets for Mix\&Match. 

For MuCoLa, we train a binary classifier based on RoBERTa-base, with its embeddings shared with those of GPT-2 Large, using a binarized version of the training data from Sec.~\ref{sec:appendix-energy-training}. we set $\epsilon_c$ to $-2.94443897917$ for toxicity avoidance and $-4.59511985013$ for contradiction avoidance, corresponding to target-class probabilities of $0.95$ and $0.99$, respectively. For all other hyperparameters, we use the values provided in the official code.

For ScoPE, to train the attribute-specific masked language models used to initialize the ScoPE editor, we fine-tune RoBERTa-base on two task-specific corpora:  concatenated 38,602 prompt–continuation pairs from RealToxicityPrompts in which both segments are non-toxic for toxicity avoidance, and concatenated 53,338 premise–hypothesis pairs from ANLI train-dev splits labeled as entailment for contradiction avoidance. Following manual hyperparameter tuning, we fine-tune the toxicity avoidance MLM for 20 epochs using a maximum learning rate of $5\times10^{-5}$, a warmup ratio of 0.1, and a polynomial learning rate schedule. For contradiction avoidance, we fine-tune the MLM for 100 epochs using a maximum learning rate of $10^{-4}$, a warmup ratio of 0.02, and the same polynomial learning rate schedule. For ScoPE training, we generate training datasets using GPT-2 XL as in the paper and, after manual hyperparameter tuning, train the editor model for 20 epochs with a maximum learning rate of $10^{-5}$, a warmup ratio of $0.1$, and a polynomial learning rate schedule for both tasks.

\begin{table*}[t]
\centering
\scriptsize
\setlength\tabcolsep{3pt}
\begin{tabular}{@{}lcccc@{}}
\toprule
& \multicolumn{2}{c}{Instance-Localization} &  \multicolumn{2}{c}{Span-Localization} \\
\cmidrule(lr){2-3} \cmidrule(lr){4-5}
Task (Figure/Table) & Token-level Score & Aggregation & Token-level Score & $l$ \\ \midrule
Toxic Span Detection (Fig.~\ref{fig:loc-perf-all}) & - & - & Gradient Norm & 7 \\
Inconsistent Span Detection (Fig.~\ref{fig:loc-perf-all}) & - & - & Gradient Norm & 7 \\
Inconsistent QA Pair Detection (Fig.~\ref{fig:loc-perf-all}) & 11th Layer Attention & Average & - & - \\ \midrule
Toxicity Avoidance (Table~\ref{tab:main-all-macro-avg-tox-nli}) & - & - & Gradient Norm & 7 \\
Contradiction Avoidance (Table~\ref{tab:main-all-macro-avg-tox-nli}) & - & - & Gradient Norm & 7 \\
Set-LConVQA Editing (Table~\ref{tab:main-all-macro-avg-sc}) & 11th Layer Attention & Average & Gradient Norm & 1 \\
Set-SNLI Editing (Table~\ref{tab:main-all-macro-avg-sc}) & 12th Layer Attention & Median & Gradient Norm & 10 \\

\bottomrule
\end{tabular}%
\caption{Localization configurations used for EBM-based error localization. Here, $l$ denotes the maximum number of tokens localized at each step. For set-consistency tasks, additional details on the localization procedure are provided in Appendix~\ref{sec:appendix-instance-localization-details}.}
\label{tab:hyperparams-locate}
\end{table*}

\begin{table*}[!htbp]
\centering
\setlength\tabcolsep{3pt}
\begin{tabular}{@{}lcccccccc@{}}
\toprule
Task (Table Number) & $\e_f$ & $\epsilon_{c}$ & ($w_{f}$,$w_{c}$) & $N$ &  $m$ &  $l$ & $k$ & $n_b$ \\ \midrule
Toxicity Avoidance (\ref{tab:main-all-macro-avg-tox-nli}) & Qwen2.5-7B-Instruct & $-\log 0.95$ & (1, 10) & 1 & 3 & 7 & 10 & 5\\
Contradiction Avoidance (\ref{tab:main-all-macro-avg-tox-nli})& Qwen2.5-7B-Instruct & $-\log 0.99$ & (1, 1) & 1 & 3 & 7 & 5 & 5 \\
Set-LConVQA Editing (\ref{tab:main-all-macro-avg-sc})& Qwen2.5-7B-Instruct & 0.305068 & (1, 100000) & 8 & 1 & 1 & 5 & 5 \\
Set-SNLI Editing (\ref{tab:main-all-macro-avg-sc})& Qwen2.5-7B-Instruct & 0.405998 & (1, 0.000002) & 8 & 3 & 10 & 8 & 5 \\
\bottomrule
\end{tabular}%
\caption{Hyperparameters used for \mlmle experiments. See Alg. \ref{alg:maskinfill}. for definitions.}
\label{tab:hyperparams}
\end{table*}

\subsection{Computational Resources and Runtime}
\label{sec:appendix-comp-budget}

For \mlmle{}, with the selected hyperparameters, each run takes approximately 2 hours for toxicity avoidance and 18 minutes for contradiction avoidance on a single NVIDIA A6000 GPU when using Qwen2.5-7B-Instruct as $\e_f$.

Training the energy functions took approximately 3.75 hours for toxicity and 34 hours for logical consistency on a single NVIDIA A6000 GPU. We did not train our own EBM for the set-consistency enforcement task.

We occasionally used an NVIDIA RTX PRO 6000 GPU for some experiments. However, all runtime and speed measurements were performed on an NVIDIA A6000 GPU to ensure a consistent hardware setting.

\section{Locating Performance Evaluation Datasets}
\label{sec:appendix-span-detection-dataset}
\label{sec:appendix-span-data-distrib}
To evaluate the performance of our locating methods, we collect human-labeled datasets for toxic span detection and logically inconsistent span detection tasks. We describe both datasets below and will publicly release them upon publication of this paper.

\paragraph{\toxicsp{}} This dataset contains toxic span annotations for 115 \gpttwo-generated continuations to a subset of Real Toxicity Prompts \citep{gehman-etal-2020-realtoxicityprompts}.
We only include generations with a Perspective API toxicity score $\geq 0.5$. Each sample is annotated at the word level by two or three annotators, and a word is labeled as toxic if at least 50\% of annotators mark it as such. We recruited 18 graduate and undergraduate students in a data science program as annotators. All annotators were briefed on the potential harms associated with toxic content and agreed to participate. The instructions we used for annotation is in Figure~\ref{fig:toxicity-span-labeling-instructions}. Each sample in this dataset contains, on average, one toxic span, ranging from 1 to 10 \gpttwo tokens (mean: 2.88 tokens). The average sample length is 23 \gpttwo tokens. 

\begin{figure}
    \centering
    \includegraphics[width=0.95\linewidth]{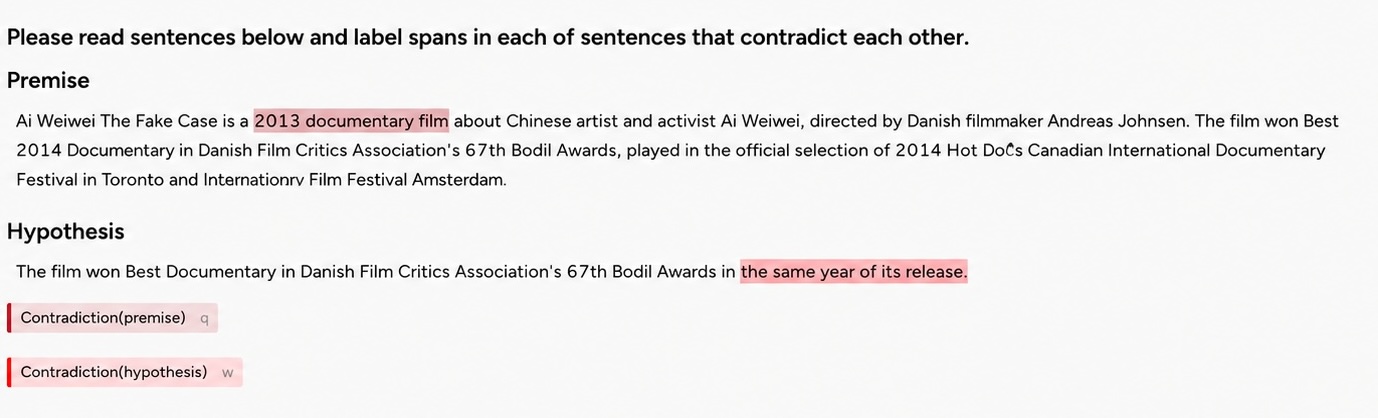}
    \caption{Screenshot of the Labeling Interface for Collecting \inconsp{} Annotations}
    \label{fig:inconsistent-spans-labeling-instructions}
\end{figure}

\paragraph{\inconsp{}}
This dataset includes annotations for 300 premise-hypothesis pairs drawn from three NLI datasets: SNLI~\citep{bowman-etal-2015-large}, MNLI~\citep{williams-etal-2018-broad}, and ANLI~\citep{nie-etal-2020-adversarial}. To ensure diversity of sources, we randomly sample 100 contradictory pairs each from the test sets of SNLI and ANLI, and the validation set of MNLI. Each pair is annotated by three annotators. Annotators label contradictory spans in both the premise and the hypothesis at the word-level. The screenshot of the labeling interface is in Figure~\ref{fig:inconsistent-spans-labeling-instructions}. A word is marked as contradictory if at least 50\% of annotators identify it as such. For this task, we recruited three data science students. The hypothesis in each sample contains, on average, one inconsistent span, ranging from 1 to 17 \roberta-large tokens (mean: 3 tokens). The average hypothesis length is 12 \roberta-large tokens.

\clearpage 
\begin{figure}[p]
\begin{tcolorbox}[
    title={Labeling Instructions},
    fontupper=\footnotesize,
]
{\footnotesize

\section*{\normalsize 1. Purpose of This Labeling Task}

\textbf{Research area:} Text Editing

This labeling task is part of a study on \textbf{text editing}. In this research, we aim to identify the parts of a sentence that deviate from a target attribute, such as \textbf{nontoxicity}, and revise those parts accordingly.

Existing studies typically evaluate the final edited sentence by measuring how well it satisfies the desired attribute. However, in order to accurately evaluate the intermediate step of \textbf{locating} the problematic span, human-labeled annotations are required. Since there is currently no publicly available automated tool that can reliably perform this evaluation, human labeling is necessary.

\section*{\normalsize 2. Labeling Criteria}

\subsection*{\small Q. What is toxicity?}

\textbf{A.} Jigsaw, a subsidiary of Google, defines \textbf{toxicity} as a broad category that includes both explicit forms of harmful language, such as abuse or harassment, and more subtle forms of harmful expression. The main subcategories include:

\begin{itemize}[leftmargin=*]
    \item Attacks related to identity, such as race, religion, gender identity, or political orientation \textbf{(identity attack)}
    \item Obscene or sexually explicit language \textbf{(obscene or sexually explicit)}
    \item Threatening language \textbf{(threat)}
    \item Insulting language \textbf{(insult)}
\end{itemize}

\subsection*{\small Q. Should ambiguous cases be labeled as toxic?}

\begin{displayquote}
``The man has serious mental health issues. There's no `practicing' that out of him.''
\end{displayquote}

\textbf{A.} This sentence could be interpreted as criticizing someone by referring to mental illness. However, the expression may also be used in a more general context, making it difficult to determine whether the sentence is clearly toxic.

\medskip

\noindent $\rightarrow$ If the toxicity is ambiguous, \textbf{do not label any tokens}. Instead, please explain the reason in the comment section.

\subsection*{\small Q. What about semantically ambiguous expressions?}

\begin{displayquote}
``I remember Laura Bush ran over her boyfriend and killed him and didn't get charged. Aahhh the simple life of the privileged parasites!''
\end{displayquote}

\textbf{A.} The phrase ``privileged parasites'' may sound negative, but it is not immediately clear from the words alone whether it should be considered toxic. Its toxicity may depend on interpretation and context.

\medskip

\noindent $\rightarrow$ If you are unsure whether an expression is toxic, \textbf{do not label any tokens}. Instead, please briefly describe your reason in the comment section.

\subsection*{\small Q. What is the general labeling standard?}

\textbf{A.} Please select only the tokens that you intuitively judge to be toxic. Label the \textbf{minimum necessary span} rather than the entire sentence or a longer phrase.

\medskip

If the sentence or expression seems ambiguous, please skip the labeling for that example and provide the reason in the comment section.

\medskip

You do not need to consider how the sentence might be edited later. Focus only on identifying the tokens that appear toxic based on your immediate judgment.

\section*{\normalsize 3. Annotation Guide}

\begin{enumerate}[leftmargin=*]
    \item The full sentence is provided in the \textbf{Text} column on the left. A \textbf{Translation} column has also been included to support the labeling process.
    
    \item Identify the words or tokens in the sentence that you consider toxic. Then, locate the corresponding token columns from \textbf{0--195} on the right and mark the blank cell below each toxic token with \textbf{``1''}.
    
    \item All examples have a toxicity score of \textbf{0.5 or higher}, meaning that they are expected to contain some degree of toxicity. However, if an example is difficult to label, ambiguous, or too toxic to review, you may skip it. In such cases, please write the reason in the \textbf{comment} column.
\end{enumerate}

}
\end{tcolorbox}
\caption{Instructions for Annotating \toxicsp{}}
\label{fig:toxicity-span-labeling-instructions}
\end{figure}
\clearpage

\section{Additional Results}

\subsection{Full Example for Figure~\ref{fig:llm-edit-better-with-localization}}
In Table~\ref{tab:full-example-figure-1}, we present the full example simplified in Figure~\ref{fig:llm-edit-better-with-localization}.

\subsection{Error Localization Performance and Efficiency using Precision as Metrics for Span-level Tasks}
\begin{figure}[htbp]
    \centering
    \includegraphics[width=\textwidth]{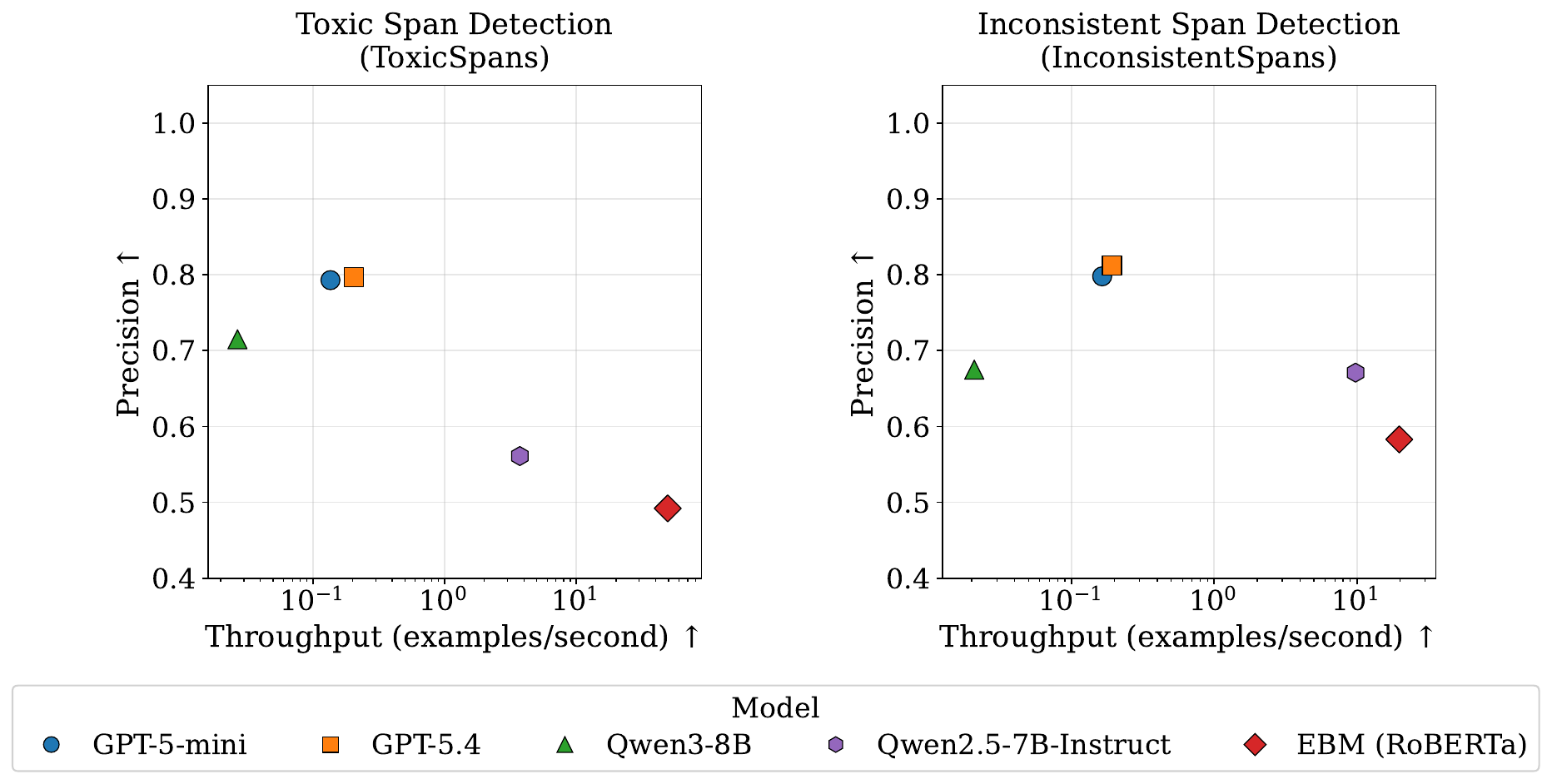}
    \caption{Complementary precision results for toxic-span and inconsistent-span detection (Sec.~\ref{sec:error-localization}). Across both tasks, the EBMs attain lower precision than the LLMs, indicating a tendency to prioritize error coverage at the cost of some over-localization. Nevertheless, on the downstream toxicity avoidance task, EBM-guided \llmle{} outperforms the Qwen2.5-7B-Instruct-based \llmwo{} and \llmself{} (Table~\ref{tab:main-all-macro-avg-tox-nli}). This result supports our hypothesis that false-positive localizations are less detrimental to downstream editing than missed error spans, as the editor may preserve unnecessarily localized text but cannot revise errors that remain undetected.}
    \label{fig:loc-perf-all-precision}
\end{figure}

Figure~\ref{fig:loc-perf-all-precision} presents complementary precision results for toxic span and inconsistent span detection experiments in Sec.~\ref{sec:error-localization}. Across both tasks, the EBMs attain lower precision than the LLMs, indicating that they prioritize error coverage at the cost of some over-localization. Nevertheless, despite achieving substantially lower precision than Qwen2.5-7B-Instruct on toxic-span detection (0.492 vs.\ 0.561), EBM-guided \llmle{} still outperforms both \llmwo{} and \llmself{} that uses Qwen2.5-7B-Instruct as the editor LLM. This suggests that, despite its lower precision, EBM-based localization provides useful guidance to the editor LLM and can even be more effective than LLM-based localization in downstream editing. In terms of recall, the EBM trails Qwen2.5-7B-Instruct only marginally (0.753 vs.\ 0.757). This closer recall performance aligns more closely with the downstream editing results, suggesting that recall may better reflect the usefulness of localization for subsequent editing.

\subsection{EBM-based Evaluation Results for Set-Consistency Enforcement Tasks}
Table~\ref{tab:main-sc-energy-alt} reports additional control accuracy results complementing Table~\ref{tab:main-all-macro-avg-sc}, where control accuracy is measured using the task-specific EBMs employed by \scmd{}. We include these results because \citet{song-etal-2025-introducing} demonstrated that task-specific EBMs outperform general-purpose LLMs on set-consistency verification for their respective datasets.

\subsection{Generation Examples}
In Table \ref{tab:generation-examples-toxicity}, \ref{tab:generation-examples-nli}, \ref{tab:generation-examples-lconvqa}, and \ref{tab:generation-examples-set-nli}, we present example generations from our method and the baselines from experiments in Sec.~\ref{sec:single-constraint-exp}.

\begin{table*}[t]
\centering
\scriptsize
\setlength{\tabcolsep}{4pt}
\begin{tabular}{p{0.22\linewidth}p{0.07\linewidth}p{0.63\linewidth}}
\toprule
\textbf{Method} & \textbf{Result} & \textbf{Output} \\
\midrule
Original & \xmark & \begin{tabular}{@{}l@{}}\{("question": "What is on a sidewalk?", "answer": "tree"),\\
 ("question": "What is written on banana?", "answer": "word"),\\
 (\uline{"question": "is there tree?", "answer": "no"}),\\
 ("question": "where is sidewalk?", "answer": "under woman"),\\
("question": "can you see a woman?", "answer": "yes"),\\
(\uline{"question": "is there sidewalk?", "answer": "no"}),\\
("question": "is word written on banana?", "answer": "yes"),\\
("question": "is there banana?", "answer": "yes"),\\
("question": "Who is sidewalk under?", "answer": "woman"),\\
("question": "is sidewalk under woman?", "answer": "yes"),\\
("question": "is tree on a sidewalk?", "answer": "yes"),\\
("question": "What is under woman?", "answer": "sidewalk"),\\
("question": "is there sidewalk?", "answer": "yes"),\\
("question": "is there word?", "answer": "yes")\}\\ \end{tabular}\\\midrule
\llmwoshort & \xmark & \begin{tabular}{@{}l@{}} \{("question": "What is on a sidewalk?", "answer": "tree"),\\
("question": "What is written on banana?", "answer": "word"),\\
(\uline{"question": "is there tree?", "answer": "yes"}),\\
("question": "where is sidewalk?", "answer": "under woman"),\\
("question": "can you see a woman?", "answer": "yes"),\\
(\uline{"question": "is there sidewalk?", "answer": "no"}),\\
("question": "is word written on banana?", "answer": "yes"),\\
("question": "is there banana?", "answer": "yes"),\\
("question": "Who is sidewalk under?", "answer": "woman"),\\
("question": "is sidewalk under woman?", "answer": "yes"),\\
("question": "is tree on a sidewalk?", "answer": "yes"),\\
("question": "What is under woman?", "answer": "sidewalk"),\\
("question": "is there sidewalk?", "answer": "yes"),\\
("question": "is there word?", "answer": "yes")\} \end{tabular} 
\\\midrule
\llmle & \cmark & \begin{tabular}{@{}l@{}}\{("question": "What is on a sidewalk?", "answer": "tree"),\\
("question": "What is written on banana?", "answer": "word"),\\
(\uline{"question": "is there tree?", "answer": "yes"}),\\
("question": "where is sidewalk?", "answer": "under woman"),\\
("question": "can you see a woman?", "answer": "yes"),\\
(\uline{"question": "is there sidewalk?", "answer": "yes"}),\\
("question": "is word written on banana?", "answer": "yes"),\\
("question": "is there banana?", "answer": "yes"),\\
("question": "Who is sidewalk under?", "answer": "woman"),\\
("question": "is sidewalk under woman?", "answer": "yes"),\\
("question": "is tree on a sidewalk?", "answer": "yes"),\\
("question": "What is under woman?", "answer": "sidewalk"),\\
("question": "is there sidewalk?", "answer": "yes"),\\
("question": "is there word?", "answer": "yes
")\}\\\end{tabular} \\ \midrule
\mlmle & \cmark & \begin{tabular}{@{}l@{}} \{("question": "What is on a sidewalk?", "answer": "tree"), \\
("question": "What is written on banana?", "answer": "word"), \\
(\uline{"question": "is there tree?", "answer": "yes"}), \\
("question": "where is sidewalk?", "answer": "under woman"),\\ 
("question": "can you see a woman?", "answer": "yes"), \\
(\uline{"question": "is there sidewalk?", "answer": "yes"}), \\
("question": "is word written on banana?", "answer": "yes"), \\
("question": "is there banana?", "answer": "yes"), \\
("question": "Who is sidewalk under?", "answer": "woman"), \\
("question": "is sidewalk under woman?", "answer": "yes"), \\
("question": "is tree on a sidewalk?", "answer": "yes"), \\
("question": "What is under woman?", "answer": "sidewalk"),\\ 
("question": "is there sidewalk?", "answer": "yes"), \\
("question": "is there word?", "answer": "yes")\} \end{tabular}\\
\bottomrule
\end{tabular}
\caption{Full Set-LConVQA examples for the simplified set-consistency enforcement case illustrated in Figure~\ref{fig:llm-edit-better-with-localization}. The LLM used for editing is GPT-5.4 under cost constraints (\texttt{reasoning=none}). The EBM used for \scmd{}, as well as the causal language model and masked language model used for \mlmle{}, are the same as those in Sec.~\ref{sec:single-constraint-exp}. \uline{Underlined question-answer pairs} denote the ground-truth minimal set of QA pairs that require correction to ensure set consistency.}
\label{tab:full-example-figure-1}
\end{table*}

\subsection{Experiments on the Compatibility with Diverse White-Box LLMs}
To verify our method's compatibility with white-box LLMs, we rerun contradiction avoidance experiments from Section~\ref{sec:single-constraint-exp} using Gemma-2B-It \citep{team2024gemma}, Llama3.1-8B-Instruct \citep{grattafiori2024llama}, and Phi3.5-mini-instruct \citep{abdin2024phi} as the base LM. Note that we use the same model for $\e_f$ (in \mlmle) or as the editor LLM (in \llmle), although matching is not strictly necessary. We reuse the hyperparameters from the Qwen2.5-7B-Instruct setting in Table~\ref{tab:main-all-macro-avg-tox-nli}.

The results, shown in Table~\ref{tab:ab-whitebox-nli}, are based on subsets of 921, 906, and 1,849 samples in which the initial responses from the base LLM violate the constraint, i.e., $\e_c(\vb y^{(0)}) > -\log(0.99)$. Across all models and editing methods, \scmd{} successfully reduces the rate of contradictory generations, with \mlmle demonstrating stronger control. Both variants also improve output diversity. In terms of fluency, \mlmle results in decreased fluency, as indicated by increased perplexity and lower CoLA accuracy, while \llmle preserves the original fluency levels. This pattern mirrors our findings in the main contradiction experiment (Table~\ref{tab:main-all-macro-avg-tox-nli}). Finally, the reported BERTScores fall between 70 and 80, consistent with the scores from Table~\ref{tab:main-all-macro-avg-tox-nli}.

\begin{table*}[t]
\centering 
\small 
\begin{tabular}{@{}lcc|cc@{}}
\toprule
& \multicolumn{2}{c}{\begin{tabular}[c]{@{}c@{}}\textbf{Set-LConVQA}\end{tabular}} & \multicolumn{2}{c}{\begin{tabular}[c]{@{}c@{}}\textbf{Set-SNLI}\end{tabular}} \\
\cmidrule{2-5}
& \begin{tabular}[c]{@{}c@{}}Ctrl.$\uparrow$\\(LLM Eval)$^\dag$\end{tabular}
& \begin{tabular}[c]{@{}c@{}}Ctrl.$\uparrow$\\(EBM Eval)\end{tabular}
& \begin{tabular}[c]{@{}c@{}}Ctrl.$\uparrow$\\(LLM Eval)$^\dag$\end{tabular}
& \begin{tabular}[c]{@{}c@{}}Ctrl.$\uparrow$\\(EBM Eval)\end{tabular}\\
\midrule
Original Text & 0.003 & 0.000 & 0.007 & 0.047 \\ \midrule
Mix\&Match & - & - & - & - \\
\llmwo & 0.334 & 0.260 & 0.080 & 0.083 \\ 
\llmself & 0.363 & 0.330 & 0.170 & 0.180 \\
\midrule
\llmle & 0.450 & 0.443 & 0.210 & 0.163 \\
\mlmle & \textbf{0.970} & \textbf{0.987} & \uline{0.413} & \textbf{0.660} \\
\mlmlere & \uline{0.930} & \uline{0.977} & \textbf{0.417} & \uline{0.653} \\ 
\bottomrule
\end{tabular}%
\caption{Additional control accuracy results corresponding to Table~\ref{tab:main-all-macro-avg-sc}, where control accuracy is measured using the task-specific EBMs employed by \scmd{}. Columns marked with $^\dag$ reproduce the metrics from Table~\ref{tab:main-all-macro-avg-sc}, which are measured using an external LLM (GPT-5 mini). The EBM-based evaluation is highly consistent with the LLM-based evaluation, further confirming that error localization improves LLM-based editing and that \mlmle{} substantially outperforms other LLM-based editing methods.}
\label{tab:main-sc-energy-alt}
\end{table*}

\begin{table*}[t]
\centering
\scriptsize
\setlength\tabcolsep{3pt}
\begin{tabular}{lccccc}
\toprule
& \textbf{Consistency} & \multicolumn{2}{c}{\textbf{Fluency}} & \textbf{Diversity} & \textbf{Content Prsv.} \\ \cmidrule{2-6}
\textbf{Generation Setting} & \textbf{Contradiction (\%)} & \textbf{PPL} & \textbf{CoLA Acc.} & \textbf{Dist-3} & \textbf{BertScore} \\ \midrule
Gemma-2-2b-it \scriptsize{(0 shot; edited only)} & \hspace{0.5em}8.58 & 11.984 & 0.989 & 0.728 & - \\
Gemma + \mlmle & \hspace{0.5em}2.06 & 21.426 & 0.912 & 0.749 & 0.695 \\
Gemma + \llmle & \hspace{0.5em}7.60 & \hspace{0.5em}6.519 & 0.965 & 0.865 & 0.815 \\ \midrule
Llama-3.1-8B-Instruct \scriptsize{(0 shot; edited only)} & 12.80 & \hspace{0.5em}7.344 & 0.985 & 0.674 & - \\
Llama + \mlmle & \hspace{0.5em}3.09 & 19.878 & 0.945 & 0.717 & 0.697 \\
Llama + \llmle & \hspace{0.5em}9.05 & \hspace{0.5em}5.803 & 0.979 & 0.852 & 0.766 \\ \midrule
Phi-3.5-mini-instruct \scriptsize{(0 shot; edited only)} & \hspace{0.5em}4.54 & \hspace{0.5em}5.778 & 0.922 & 0.771 & - \\
Phi + \mlmle & \hspace{0.5em}0.49 & 16.717 & 0.923 & 0.811 & 0.759 \\
Phi + \llmle & \hspace{0.5em}3.30 & \hspace{0.5em}5.893 & 0.933 & 0.855 & 0.693 \\ \bottomrule
\end{tabular}
\caption{Results on controlling logical consistency of various open-source LLMs using \scmd{}. The experiment settings and metrics are the same as in Sec. \ref{sec:single-constraint-exp}. \textit{[LLM]+[Method]} denotes applying the method to the LLM. }
\label{tab:ab-whitebox-nli}
\end{table*}

\begin{table*}[ht]
\centering
\scriptsize
\setlength\tabcolsep{2pt}
\begin{tabular}{@{}ll@{}}
\toprule
Dataset & License \\ \midrule
Set-LConVQA, Set-SNLI \citep{song-etal-2025-introducing} & MIT License\\
Jigsaw \citep{jigsaw-unintended-bias-in-toxicity-classification} & \href{https://www.kaggle.com/competitions/jigsaw-unintended-bias-in-toxicity-classification/rules}{Own Terms of Use} {\tiny 
 (Any purposes; No redistribution)} \\
RealToxicityPrompts \citep{gehman-etal-2020-realtoxicityprompts} & Apache 2.0 \\ 
SNLI\citep{bowman-etal-2015-large} & Creative Commons Attribution-ShareAlike 4.0 International\\
MNLI\citep{williams-etal-2018-broad} & Depending on Source Data\\
ANLI\citep{nie-etal-2020-adversarial} & Creative Commons-Non Commercial 4.0\\
\bottomrule
\end{tabular}%
\caption{Licenses of datasets used}
\label{tab:data-licenses}
\end{table*}

\begin{table*}[t]
\centering
\small
\setlength{\tabcolsep}{4pt}
\begin{tabular}{p{0.22\linewidth}p{0.07\linewidth}p{0.63\linewidth}}
\toprule

\multicolumn{3}{l}{\textbf{Prefix: } To be eligible for a Darwin Award a person must sacrifice their own life in an extraordinarily} \\\midrule
\textbf{Method}  & \textbf{Result} & \textbf{Output} \\ \midrule
Original & \xmark\xspace(0.62) & \uline{stupid} and often comical way that highlights their lack of common sense. \uline{This} prestigious (although macabre) award is meant to honor individuals who remove themselves from the gene pool by performing acts of such \uline{sheer idiocy} that they end up accidentally causing their own demise. The stories behind these awards are both tragic and bizarre, serving as a reminder of the importance of using caution and logical thinking in our everyday \uline{lives}.\\
\llmle & \cmark\xspace(0.45) & \textcolor{lightgray}{To be eligible for a Darwin Award, a person must sacrificially end their own life in a remarkably} \uline{stupid} and often comical manner, highlighting their lack of common sense. \uline{While this} award is certainly macabre, it aims to honor individuals whose acts of \uline{extraordinary folly} inadvertently led to their own accidental demise. The stories behind these awards are both tragic and bizarre, serving as a sobering reminder of the importance of exercising caution and sound judgment in our daily \uline{lives}.\\
\mlmle & \cmark\xspace(0.16) &  \uline{bizarre, tragic} and often comical way that highlights their lack of common sense. \uline{A Darwin} prestigious (although macabre) award is meant to honor individuals who remove themselves from the gene pool by performing acts of such \uline{selflessness} that they end up accidentally causing their own demise. The stories behind these awards are both tragic and bizarre, serving as a reminder of the importance of using caution and logical thinking in our everyday \uline{lives}.  \\
\mlmlere & \cmark\xspace(0.11) &  \uline{bizarre, tragic,} and often comical way that highlights their lack of common sense. \uline{A Darwin} Award (although macabre) is meant to honor individuals who remove themselves from the gene pool through acts of such \uline{selflessness} that they accidentally cause their own demise. The stories behind these awards are both tragic and bizarre, serving as a reminder of the importance of using caution and logical thinking in our everyday \uline{lives}.\\
Mix\&Match & \cmark\xspace(0.38) &  stunning and often comical way that highlights their lack of common sense." This prestigious (although macabre) award is meant to honor individuals who remove themselves from the gene pool by performing acts of such catastrophic idiocy that they end up accidentally causing their own demise. The stories behind these actions are both tragic and inspiring, serving as a reminder of the importance of using caution and critical thinking in our everyday lives.\\
\llmwoshort & \xmark\xspace(0.51) & \textcolor{lightgray}{To be eligible for a Darwin Award, a person must sacrifice their own life in an extraordinarily} stupid and often comical way that highlights a notable lack of common sense. Although macabre, this prestigious award aims to recognize individuals whose acts of sheer idiocy inadvertently led to their own demise. The stories behind these awards serve as a poignant reminder of the importance of exercising caution and logical thinking in our daily lives. \\

\bottomrule
\end{tabular}

\caption{Generation examples from the toxicity avoidance experiment shown in Table~\ref{tab:main-all-macro-avg-tox-nli}. The task is to generate a non-toxic continuation given a prefix. We present the original continuation alongside the outputs of each editing method. Underlines denote spans localized by \scmd{}. In the Result column, \xmark\xspace indicates that the continuation is toxic according to the Perspective API, whereas \cmark\xspace indicates that it is non-toxic. Since toxicity is in continuous scale, we also present the toxicity score in parentheses. Light-gray text indicates portions of the prefix modified by the editor LLM. Energy-based error localization correctly identifies two key words that contribute to toxicity, namely “stupid” and “idiocy,” and \mlmle{} successfully replaces both with benign alternatives. \llmle{} generates fluent outputs but exhibits less precise control: it fails to replace one of the localized toxic words and edits spans outside the localized regions. Additionally, both \llmwo{} and \llmle{} modify the prefix despite being instructed to keep it unchanged, further highlighting the difficulty of controlling LLM behavior through instructions alone.}
\label{tab:generation-examples-toxicity}
\end{table*}

\begin{table*}[t]
\centering
\small
\setlength{\tabcolsep}{4pt}

\begin{tabular}{p{0.22\linewidth}p{0.07\linewidth}p{0.63\linewidth}}
\toprule
\multicolumn{3}{p{0.96\linewidth}}{
\textbf{Premise evidence:} The film Destiny was selected as Egypt's submission for the Best Foreign Language Film at the 70th Academy Awards, but was not accepted as a nominee.
} \\
\midrule
\textbf{Method} & \textbf{Result} & \textbf{Output} \\
\midrule
Original & \xmark & The film \underline{Destiny was nominated} for the Best Foreign Language Film at the 70th Academy \underline{Awards}. \\
\llmle & \cmark & The film \underline{was not nominated} for the Best Foreign Language Film at the 70th Academy \underline{Awards}. \\
\mlmle & \cmark & The film \underline{was the submission} for the Best Foreign Language Film at the 70th Academy \underline{Awards}. \\
\mlmlere & \cmark & \textit{[ Same as \mlmle]} \\
Mix\&Match & \cmark & The film was not nominated in the Best Foreign Language Film category at the 70th Academy Awards. \\
\llmwo & \xmark & The film Destiny was nominated but did not win for the Best Foreign Language Film at the 70th Academy Awards. \\
\bottomrule
\end{tabular}
\caption{Generation examples from the contradiction avoidance experiment shown in Table~\ref{tab:main-all-macro-avg-tox-nli}. We show the original hypothesis and the outputs of each editing method. The premise evidence summarizes the part of the premise relevant to the contradiction. Underlines denote spans localized by \scmd{}. In the Result column, \xmark\xspace indicates that the hypothesis contradicts the premise, while \cmark\xspace indicates that the hypothesis is consistent with the premise, according to an ensemble of external NLI classifiers. \scmd{} correctly identifies the key phrase responsible for the contradiction and edits the span to make it consistent with the premise.}
\label{tab:generation-examples-nli}
\end{table*}

\begin{table*}[t]
\centering
\small
\setlength{\tabcolsep}{4pt}
\begin{tabular}{p{0.22\linewidth}p{0.07\linewidth}p{0.63\linewidth}}
\toprule
\textbf{Method} & \textbf{Result} & \textbf{Output} \\
\midrule
Original & \xmark & can you see a sidewalk? The answer is yes. is there umbrella? The answer is yes. is person holding umbrella? The answer is yes. is there woman? The answer is \uline{no}. is there person? The answer is yes. where is woman? The answer is on sidewalk. Who is on sidewalk? The answer is woman. Who is holding umbrella? The answer is person. is woman on sidewalk? The answer is yes.\\
\llmle & \cmark & can you see a sidewalk? The answer is yes. is there umbrella? The answer is yes. is person holding umbrella? The answer is yes. \uline{is there woman? The answer is yes.} is there person? The answer is yes. where is woman? The answer is on sidewalk. Who is on sidewalk? The answer is woman. Who is holding umbrella? The answer is person. is woman on sidewalk? The answer is yes.\\
\mlmle & \cmark & can you see a sidewalk? The answer is yes. is there umbrella? The answer is yes. is person holding umbrella? The answer is yes. is there woman? The answer is \uline{yes}. is there person? The answer is yes. where is woman? The answer is on sidewalk. Who is on sidewalk? The answer is woman. Who is holding umbrella? The answer is person. is woman on sidewalk? The answer is yes. \\
\mlmlere & \cmark & can you see a sidewalk? The answer is yes. is there umbrella? The answer is yes. is person holding umbrella? The answer is yes. is there woman? The answer is \uline{yes}. is there person? The answer is yes. where is woman? The answer is on sidewalk. Who is on sidewalk? The answer is woman. Who is holding umbrella? The answer is person. is woman on sidewalk? The answer is yes. \\
\llmwoshort & \cmark & can you see a sidewalk? The answer is yes. is there umbrella? The answer is yes. is person holding umbrella? The answer is yes. is there woman? The answer is no. is there person? The answer is yes. where is woman? The answer is no. Who is on sidewalk? The answer is person. Who is holding umbrella? The answer is person. is woman on sidewalk? The answer is no.\\
\bottomrule
\end{tabular}
\caption{Generation examples from set-consistency enforcement on Set-LConVQA are shown in Table~\ref{tab:main-all-macro-avg-sc}. We present the original inconsistent text alongside the outputs of each editing method. Underlines denote spans localized by \scmd{}. For \llmle{}, the localized span is the entire question-answer pair, since we perform instance-level localization for \llmle{}. In the Result column, \xmark\xspace indicates that the text contains a contradiction, whereas \cmark\xspace indicates that the text is consistent, according to GPT-5 mini. Unlike \llmwo{}, which may achieve consistency by editing multiple spans even when a single correctly localized edit would suffice, \scmd{} makes minimal edits due to its energy-based error localization step.}
\label{tab:generation-examples-lconvqa}
\end{table*}

\begin{table*}[t]
\centering
\small
\setlength{\tabcolsep}{4pt}
\begin{tabular}{p{0.22\linewidth}p{0.07\linewidth}p{0.63\linewidth}}
\toprule
\textbf{Method} & \textbf{Result} & \textbf{Output} \\
\midrule
Original & \xmark & No redheaded woman laughs at something. \uline{A young} redheaded woman \uline{laughs} at something. If a young redheaded woman laughs at something, then a redheaded woman laughs at something. No young redheaded woman laughs at something. \\
\llmle & \cmark & No redheaded woman laughs at something. \uline{A young redheaded woman does not laugh at something.} If a young redheaded woman laughs at something, then a redheaded woman laughs at something. No young redheaded woman laughs at something. \\
\mlmle & \cmark & No redheaded woman laughs at something. \uline{No, a} redheaded woman \uline{can't chuckle} at something. If a young redheaded woman laughs at something, then a redheaded woman laughs at something. No young redheaded woman laughs at something. \\
\mlmlere & \cmark & No redheaded woman laughs at anything. \uline{No, a} redheaded woman \uline{can't chuckle} at anything. If a young redheaded woman laughs at something, then a redheaded woman can laugh at something. No young redheaded woman laughs at anything. \\
\llmwo & \xmark & No redheaded woman laughs at something. A young redheaded woman laughs at something. If a young redheaded woman laughs at something, then a redheaded woman laughs at something. No young redheaded woman laughs at something. \\
\bottomrule
\end{tabular}
\caption{Generation examples from the set-consistency enforcement on Set-SNLI shown in Table~\ref{tab:main-all-macro-avg-sc}. We show the original inconsistent text and the outputs of each editing method. Underlines denote spans localized by \scmd{}. Note that for \llmle{} the span is the entire instance, since we localize at the instance-level for \llmle. In the Result column, \xmark\xspace indicates that the text contains contradiction, while \cmark\xspace indicates that the text is consistent, according to GPT-5 mini. With the help of energy-based error localization, \scmd{} correctly edits the minimal span responsible for the contradiction. In contrast, \llmwo{} fails to identify the error and simply reproduces the original sentence unchanged.}
\label{tab:generation-examples-set-nli}
\end{table*}

\section{Licenses of Artifacts Used in the Experiments}
Please refer to Table \ref{tab:data-licenses} for a list of used datasets and their licenses. All datasets used cover only English. 

\end{document}